\documentclass[11pt]{article}
\usepackage[final]{acl}
\usepackage{listings}
\usepackage{times}
\usepackage{latexsym}
\usepackage{amsmath}
\usepackage{amssymb}
\usepackage{booktabs}
\usepackage{graphicx}
\usepackage{multirow}
\usepackage{microtype}
\usepackage{inconsolata}
\usepackage{enumitem}
\usepackage{verbatim}
\usepackage{xcolor}
\usepackage{colortbl}
\usepackage{arydshln}
\setlist{nosep,leftmargin=*}

\usepackage{pgfplots}
\pgfplotsset{compat=1.18}
\usetikzlibrary{patterns.meta}
\definecolor{colS}{HTML}{EB6834}
\definecolor{colF}{HTML}{2A78D6}
\definecolor{inkPrimary}{HTML}{1F1F1E}
\definecolor{inkSecondary}{HTML}{52514E}
\definecolor{gridGray}{HTML}{C9C8C3}

\definecolor{inkPrimary}{HTML}{1F1F1F}
\definecolor{inkSecondary}{HTML}{555555}
\definecolor{gridGray}{HTML}{C8C8C8}

\definecolor{cfGold}{HTML}{4A3AA7}
\definecolor{cfLure}{HTML}{E34948}
\definecolor{cfAbst}{HTML}{A9A9A9}
\definecolor{cfOther}{HTML}{DADADA}

\newcommand{\promptfield}[1]{{\normalfont\itshape\color{colF}\textlangle#1\textrangle}}
\lstdefinestyle{prompt}{basicstyle=\ttfamily\footnotesize, breaklines=true, breakatwhitespace=true, columns=fullflexible, breakindent=0pt, frame=single, rulecolor=\color{gray!50}, backgroundcolor=\color{gray!6}, framesep=5pt, xleftmargin=5pt, xrightmargin=5pt, aboveskip=0pt, belowskip=0pt, escapeinside={(*}{*)}}

\title{\textsc{MisleadQA}: Diagnosing LLM Failures under Misleading Context}
\author{
\textbf{Hui Wang}\textsuperscript{1,\dag,*} \;\;\;
\textbf{Bohao Chu}\textsuperscript{1,\dag} \;\;\;
\textbf{Qianli Wang}\textsuperscript{2} \;\;\; \\
\textbf{Shuning Zhang}\textsuperscript{3} \;\;\;
\textbf{Norbert Fuhr}\textsuperscript{1} \\
\textsuperscript{1}University of Duisburg-Essen,
\textsuperscript{2}Technische Universit\"at Berlin,
\textsuperscript{3}Tsinghua University \\
\textsuperscript{\dag}Equal contribution. \;\;\;
\textsuperscript{*}\textbf{Correspondence}: \texttt{{hw1361376751@gmail.com}}
}
\begin{document}
\maketitle

\begin{abstract}
Large language models (LLMs) are widely used to answer questions from retrieved or user-provided context. However, provided context can be factually true yet out-of-scope, describing a different entity, time, or location from the one the question targets, and still lead models to a wrong answer. Prior work has mainly studied factually false or irrelevant context. It remains unclear whether such out-of-scope context misleads models as much as false context, and whether adding gold context prevents models from being misled by either. We introduce \textbf{\textsc{MisleadQA}}, a diagnostic dataset of multiple-choice questions with an abstention option. Each question pairs gold context (G) with true but out-of-scope context (S) and false but in-scope context (F) supporting the same wrong answer. We evaluate six open-weight LLMs. On questions they answer correctly without context, providing S alone leads them to switch to this wrong answer in 29.3\% of cases, and providing F alone does so in 75.4\%. Adding G nearly removes the effect of S, but F still misleads models in 21.0\% of cases, especially when F comes first. A context-faithful instruction reduces these errors mainly through abstention and lowers accuracy even with G. These findings suggest that evaluations should check whether context applies to the question, not only whether it is true, and should report correct, misled, and abstention rates together. Data and code are available at \url{https://anonymous.4open.science/r/MisleadQA}.
\end{abstract}

\section{Introduction}
\label{sec:intro}

\begin{figure*}[t]
\centering
\includegraphics[width=\textwidth]{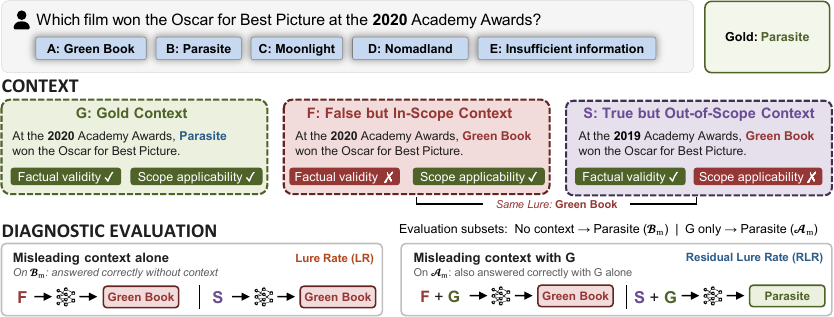}
\caption{\textsc{MisleadQA} example and evaluation. (1) Gold context (G) states the correct answer, false but in-scope context (F) makes a false claim about the target year, and true but out-of-scope context (S) states a true fact about a different year. F and S support the same lure, \emph{Green Book}. (2) Each misleading context is tested alone on $\mathcal B_m$ and together with G on $\mathcal A_m$, to measure lure rate (LR) and residual lure rate (RLR), respectively (\S\ref{sec:metrics}). Model outputs are illustrative of the main findings. The misleading-first order is shown, and both orders are evaluated (\S\ref{sec:results}).}
\label{fig:picture}
\vspace{-10pt}
\end{figure*}

Large language models (LLMs) are widely used to answer questions from context that is retrieved or user-provided \citep{NEURIPS2020_6b493230, ram-etal-2023-context, chu2026locallysoundgloballyinsufficient}. To support the correct answer, the context must meet two conditions. It must be true (\emph{factual validity}), and it must match the scope the question asks about, such as its entity, time, or location (\emph{scope applicability}). Context may meet the first condition but fail the second. For example, it may correctly state a fact for a different year than the one asked about. We call such context \emph{out-of-scope}. It points to a wrong answer without making any false claims (Figure~\ref{fig:picture}). A model must therefore check not only whether the context is true but also whether it applies to the question.

Prior work shows that models can be misled by factually false or counterfactual context \citep{longpre-etal-2021-entity,pan-etal-2023-attacking,wu2024clasheval}, and even by irrelevant context \citep{shi2023distracted}. Out-of-scope context lies between these cases: unlike false context, it is factually true, and unlike irrelevant context, it is closely related to the question. Some studies include such context \citep{chen2024rgb,amiraz-etal-2025-distracting,shafiei-etal-2026-truthtrap}, but none of them compares it with false context that supports the same wrong answer while keeping the question fixed (\S\ref{sec:related}). It therefore remains unclear whether out-of-scope context misleads models as much as false context does, and whether adding gold context protects models against either.

Answering these questions requires true context that does not apply to the question, and false context to compare it with. We therefore introduce \textsc{MisleadQA}, a dataset of 959 multiple-choice questions with an abstention option, built from Federal Reserve Economic Data (FRED)\footnote{\fontsize{8.5}{10}\selectfont\url{https://fred.stlouisfed.org/}, accessed Oct 7, 2026.}, SituatedQA \citep{zhang-choi-2021-situatedqa}, and AmbigDocs \citep{lee2024ambigdocs}. Each question has a target scope and three contexts. Gold context (G) states the correct answer. True but out-of-scope context (S) states a true fact about a different entity, time, or location, like near-miss context from retrieval. False but in-scope context (F) assigns the value from S to the target scope. S and F thus support the same wrong answer, the \emph{lure}, with everything else fixed. This lets us test whether S misleads models as much as F. Since S and F differ in both truth and scope, the comparison does not separate them.

We then conduct a diagnostic evaluation of six open-weight LLMs on \textsc{MisleadQA}, testing S and F alone and together with G, in both orders. To measure how context changes answers that models would otherwise get right, we focus on questions each model answers correctly without any context. On these questions, providing S alone leads models to switch to the lure in 29.3\% of cases, and providing F alone does so even more often, in 75.4\%. When G is added, such erroneous switches drop to 0.9\% for S but remain at 21.0\% for F.\footnote{Rates are averaged equally across models. Rates with G average both orders unless stated otherwise, and use the questions also answered correctly with G alone, which include over 99\% of the questions above (\S\ref{sec:metrics}).} Order matters mainly for F. When F is presented before G, erroneous switches rise sharply to 38.1\%, compared with only 3.9\% when G comes first.

We further analyze why S may mislead models and whether a context-faithful instruction can help. One possible reason is that models fail to notice the scope mismatch. To test this, we ask models in a separate task whether S applies to the question, using a 200-question subset. They judge only 62.4\% of S contexts as out-of-scope, even though S states its scope explicitly. However, a correct judgment does not ensure a correct answer. Even when models judge S correctly, they still choose the lure on 20.3\% of these questions when given S alone. We also test a context-faithful instruction, which tells models to use only applicable context and to abstain otherwise. It roughly halves lure selection under S, but mainly because models abstain far more often. It also lowers accuracy even when G is present. These results show that lower lure rates alone can hide a loss of correct answers, so evaluations of context-grounded question answering should report correct, misled, and abstention rates together. Overall, our contributions are:

\begin{itemize}
  \item We construct \textsc{MisleadQA}, a dataset that pairs true but out-of-scope context with matched false context, keeping the question, options, wrong answer, and gold context fixed (\S\ref{sec:dataset}).
  \item We perform a diagnostic evaluation of six open-weight LLMs (\S\ref{sec:problem}), showing that out-of-scope context alone often misleads models, and that adding gold context removes most of these errors but not those caused by false context, especially when the false context comes first (\S\ref{sec:results}).
  \item We conduct targeted analyses, showing that models often fail to recognize S as out-of-scope and can still choose the lure even when they do, and that a context-faithful instruction reduces lure selection mainly through abstention while lowering accuracy when G is present (\S\ref{sec:analysis}).
\end{itemize}

\section{Related Work}
\label{sec:related}

\textbf{Context-grounded question answering.} LLMs often answer questions using context that is either retrieved from an external source \citep{NEURIPS2020_6b493230,ram-etal-2023-context} or provided directly by the user \cite{chu-etal-2025-tracsum}. Such context may conflict with the model's parametric knowledge. Prior work studies these knowledge conflicts \citep{longpre-etal-2021-entity,xu-etal-2024-knowledge-conflicts} and encourages models to stay faithful to the given context \citep{zhou-etal-2023-context}. However, following the context is useful only if it is true and applies to the question. In practice, context can fail either condition. It can be false, due to errors or misinformation in its source. It can also be true but out-of-scope, such as a retrieved passage about a similar entity or a different year, or a user document that does not match the exact question. In both cases, a model that faithfully follows the given context would end up answering the question incorrectly. We therefore refer to both types collectively as misleading context.

\vspace{0.1cm}
\noindent\textbf{Misleading context.} Most prior work on misleading context focuses on false context, which can override a model's correct prior answer and impair question answering \citep{pan-etal-2023-attacking,hong-etal-2024-gullible,wu2024clasheval}. Other studies show that irrelevant context can distract models \citep{shi2023distracted,amiraz-etal-2025-distracting} and that passage order affects which evidence models use \citep{xie2024adaptive,tan-etal-2024-blinded}. Fewer studies examine true but out-of-scope context. RGB includes reports about other years, but tests them separately from false passages \citep{chen2024rgb}. \citet{amiraz-etal-2025-distracting} test distracting passages with a gold passage in both orders, but without a matched false passage. TruthTrap studies truthful hints rather than retrieved context \citep{shafiei-etal-2026-truthtrap}. None of these studies compares out-of-scope and false context directly. We pair the two per question so that both support the same wrong answer, and test each alone and with gold context in either order.

\vspace{0.1cm}
\noindent\textbf{Scope in question answering.} Several datasets study questions whose answers depend on scope. SituatedQA covers answers that change with time and location \citep{zhang-choi-2021-situatedqa}, AmbigDocs covers distinct entities that share a name \citep{lee2024ambigdocs}, and CondAmbigQA covers answers under explicit conditions \citep{li-etal-2025-condambigqa}. RAMDocs further combines entity ambiguity with misinformation and noise \citep{wang2025ramdocs}. These datasets test whether models can find the answer for the right scope. We instead fix the target scope and test whether misleading context pulls models away from answers they already know. We build on SituatedQA and AmbigDocs to construct out-of-scope context and its matched false context.

\section{\textsc{MisleadQA}}
\label{sec:dataset}

\textsc{MisleadQA} is a diagnostic dataset built to answer the two questions raised in \S\ref{sec:intro}: 1) whether out-of-scope context misleads models as much as false context, and 2) whether gold context prevents both errors. To this end, each item pairs a multiple-choice question with three kinds of context: gold context (G), true but out-of-scope context (S), and false but in-scope context (F). Each question has five options, including one for abstention. S and F support the same wrong answer, so their effects can be compared on the same question. The main split contains 959 items covering entity, time, and location mismatches, and a 25-item supplement on condition mismatches is analyzed separately (Table~\ref{tab:data-stats}). Below, we formalize the design (\S\ref{sec:task}), describe the sources (\S\ref{sec:sources}), and explain how we construct (\S\ref{sec:construction}) and validate (\S\ref{sec:qc}) the items.

\subsection{Dataset Design}
\label{sec:task}

\textbf{Questions and scopes.} Each item contains a question $q$ that asks for the value of an attribute $p$ (e.g., minimum wage) at a \emph{target scope} $s$ (e.g., California on January 1, 2015). A scope is a set of qualifiers, such as an entity, a time, and a location, that determine the answer. Let $v(\cdot)$ map each scope to the source-supported value of $p$ at that scope. The gold answer is $y^* = v(s)$. We also choose an \emph{alternative scope} $s'$ (e.g., California on January 1, 2016) that changes exactly one qualifier of $s$ and keeps the others fixed. Its value $\ell = v(s')$ must differ from $y^*$, and we call it the \emph{lure}. Both values must be unique at the requested precision.

\vspace{0.1cm}
\noindent\textbf{Contexts.} Each context is a single claim $c = \langle s_c, v_c \rangle$ with a stated scope $s_c$ and an asserted value $v_c$. It is \emph{true} if $v_c = v(s_c)$, and \emph{in-scope} if $s_c = s$. Each item has three contexts:
\[
\begin{aligned}
\mathrm G &= \langle s, y^* \rangle && \text{true, in-scope},\\
\mathrm S &= \langle s', \ell \rangle && \text{true, out-of-scope},\\
\mathrm F &= \langle s, \ell \rangle && \text{false, in-scope}.
\end{aligned}
\]
Scopes are compared after normalizing years, jurisdictions, and entity identifiers. We exclude any S whose stated range covers $s$, such as a time span that includes the target date.

\vspace{0.1cm}
\noindent\textbf{Design rationale.} S and F support the same lure but assign it to different scopes. They share the question, options, gold answer, and G (Figure~\ref{fig:picture}), so F serves as a matched control for S, as in counterfactual-context settings \citep{longpre-etal-2021-entity,xie2024adaptive}. The two also relate differently to G. G is consistent with S, because they concern different scopes, but it contradicts F. Adding G therefore tests whether models respond differently to the two kinds of misleading context. Since S and F differ in both truth and scope, their comparison reflects the joint effect of the two factors.

\subsection{Data Sources}
\label{sec:sources}

The main split draws on three sources, each covering one or two scope dimensions (Table~\ref{tab:data-stats}). FRED is a public database of economic time series maintained by the Federal Reserve Bank of St.~Louis. We use its 44 annual minimum-wage series, which cover 43 states and the District of Columbia from 1990 to 2024. Because each value is tied to a specific state and date, we can change the time or the location while keeping the other fixed, and we construct questions that ask for a state's basic minimum wage on a given date. SituatedQA \citep{zhang-choi-2021-situatedqa} provides natural questions whose answers depend on time or location, and we use it for time and location mismatches. AmbigDocs \citep{lee2024ambigdocs} provides Wikipedia documents about distinct entities that share a name, and we use it for entity mismatches. CondAmbigQA \citep{li-etal-2025-condambigqa} supplies the 25-item condition supplement. Items derived from the same jurisdiction or upstream question form a cluster, which we use for statistical inference (\S\ref{sec:inference}).

\begin{table}[t]
\centering
\small
\setlength{\aboverulesep}{0pt}
\setlength{\belowrulesep}{0pt}
\renewcommand{\arraystretch}{1.1}
\begin{tabular*}{\columnwidth}{@{\extracolsep{\fill}}llrr@{}}
\toprule
\textbf{Source} & \textbf{Scope} & \textbf{Items} & \textbf{Clusters}\\
\midrule
\rowcolor{gray!12}[0pt][0pt]\multicolumn{4}{@{}l@{}}{\textit{Main split}}\\
\multirow{2}{*}{FRED}       & Time     & 200 & 43\\
                            & Location & 200 & 44\\
\hdashline[1pt/1.5pt]
\multirow{2}{*}{SituatedQA} & Time     & 183 & 137\\
                            & Location & 184 & 116\\
\hdashline[1pt/1.5pt]
AmbigDocs                   & Entity   & 192 & 192\\
\hdashline[1pt/1.5pt]
Total                       &          & \textbf{959} & \textbf{489}\\
\midrule
\rowcolor{gray!12}[0pt][0pt]\multicolumn{4}{@{}l@{}}{\textit{Supplement}}\\
CondAmbigQA                 & Condition & 25 & 20\\
\bottomrule
\end{tabular*}
\caption{\textsc{MisleadQA} composition. Clusters group items that share a target jurisdiction (FRED) or an upstream question (other sources). FRED time and location items share jurisdiction clusters, so the main total counts distinct clusters. Primary analyses exclude the condition supplement.}
\label{tab:data-stats}
\vspace{-10pt}
\end{table}

\subsection{Construction}
\label{sec:construction}

We build items from all sources under the same construction guideline (Appendix~\ref{app:construction}). Each item is built in three steps: writing the question, creating the contexts, and selecting the answer options.

\vspace{0.1cm}
\noindent\textbf{Questions.} Each item starts from a fact pair: the gold fact $\langle s, y^* \rangle$ and the lure fact $\langle s', \ell \rangle$. We verify that both facts are supported by the source in their subject, attribute, scope, and value, and that $y^*$ and $\ell$ remain distinct after normalizing aliases and formats. We then write a question that specifies the target scope $s$ without revealing either value.

\vspace{0.1cm}
\noindent\textbf{Contexts.} G and S express the gold and lure facts, respectively, and state all relevant scope qualifiers explicitly. Depending on the source, they are generated from FRED records with fixed templates, taken as contiguous sentences from source documents, or written as faithful restatements of source content, with GPT-6 Astra \citep{openai2026gpt6astra} assisting in wording. F is derived from G by replacing every expression of the gold answer, including aliases, with the lure, leaving all other text unchanged. G and F therefore differ only in the asserted value. We exclude items in which this substitution leaves a gold mention, produces ungrammatical text, or creates unintended contradictions (Appendix~\ref{app:fs-similarity}).

\vspace{0.1cm}
\noindent\textbf{Answer options.} Each item has five options: the gold answer $y^*$, the lure $\ell$, two distractors $d_1$ and $d_2$, and \emph{Insufficient information} ($\bot$), which is always option E. Distractors are values of the same attribute at other documented scopes, preferably from the same upstream question. They must be wrong for the target scope, distinct from the other options after normalization, similar to the gold answer in type and precision, and absent from the question and contexts. Because models are sensitive to option positions \citep{zheng2024large}, we balance the gold and lure positions across A--D and the 12 ordered position pairs within each source--scope group of the main split, with counts differing by at most one item (Appendix~\ref{app:positions}).

\subsection{Quality Control}
\label{sec:qc}

We apply three layers of quality control: automated checks, model screening, and a human audit.

\vspace{0.1cm}
\noindent\textbf{Automated checks.} Scripts verify that each item is well-formed, that its options remain distinct after normalization, and that no distractor appears in the question or contexts. They also check that every fact is traceable to its source, that F differs from G only in the edited answer spans, and that context hashes and release checksums match. All released items pass these checks.

\vspace{0.1cm}
\noindent\textbf{Semantic screening.} Two models outside the evaluation panel screen every item with a shared prompt: GPT-6 Astra \citep{openai2026gpt6astra} in a fresh session without construction history, and Claude Opus 5.5 \citep{anthropic2026opus55}, which was not used in construction. The prompt asks whether each fact is source-supported, whether G and F match the target scope while S differs in exactly one dimension, whether the gold answer is unique, whether F reads fluently, and whether the distractors are valid (Appendix~\ref{app:screening}). Either model flags at least one criterion for 266 items. A human reviewer inspects each flagged item, excludes 64 of them, and records the reason for every decision.

\vspace{0.1cm}
\noindent\textbf{Human audit.}
\label{sec:audit}
To check whether the final items satisfy the intended design, we audit a stratified sample of 150 main items: 15 per FRED scope, 35 per SituatedQA scope, and 50 from AmbigDocs. Each item is independently reviewed by two of six annotators, with 50 items per annotator, using the criteria in Appendix~\ref{app:audit}. Both annotators judge 149 of the 150 items (99.3\%) to satisfy all criteria. All model evaluations use the same frozen version of \textsc{MisleadQA} (Appendix~\ref{app:final-release}).

\section{Experiments}
\label{sec:problem}

We evaluate how true but out-of-scope context (S) and false but in-scope context (F) affect model answers, when presented alone and together with gold context (G), on questions each model already answers correctly without any context. This section describes the evaluation setup (\S\ref{sec:conditions}), the metrics (\S\ref{sec:metrics}), and statistical inference (\S\ref{sec:inference}).

\subsection{Experimental Setup}
\label{sec:conditions}

\textbf{Task and conditions.} Given a question $q$ about a target scope, its option set $O = \{y^*, \ell, d_1, d_2, \bot\}$ (\S\ref{sec:construction}), and a possibly empty context sequence $C$ shown after the options, model $m$ predicts the option $\hat{o}$ to which it assigns the highest probability:
\[
\hat{o} = \arg\max_{o \in O}\, p_m(o \mid q, O, C).
\]
Here, $y^*$ and $\ell$ are the gold answer and the lure (\S\ref{sec:task}), $d_1$ and $d_2$ are distractors (\S\ref{sec:construction}), $\bot$ denotes \emph{Insufficient information}, and $p_m(o \mid \cdot)$ is the probability that model $m$ assigns to the letter of option $o$. The sequence $C$ consists of zero, one, or two of the contexts G, S, and F (\S\ref{sec:task}). We evaluate every item under eight context conditions:
\[
\text{\small $C \in \{\varnothing,\ \mathrm G,\ \mathrm S,\ \mathrm F,\ (\mathrm G,\mathrm S),\ (\mathrm S,\mathrm G),\ (\mathrm G,\mathrm F),\ (\mathrm F,\mathrm G)\}$,}
\]
where $\varnothing$ denotes no context and the order within parentheses is the order of presentation in the prompt. Each condition is sent as a separate request without shared conversation history, and the question text and option order are fixed across conditions, so only the context sequence $C$ changes.

\vspace{0.1cm}
\noindent\textbf{Prompt and scoring.} The prompt presents the question, the options, and any contexts, without instructing the model how to use the contexts, and asks for a single option letter. We classify each response as \emph{gold} ($y^*$), \emph{lure} ($\ell$), \emph{abstain} ($\bot$),  or \emph{other} (either distractor). All answers are scored against the target-scope gold answer, even when a context supports another option. All four categories remain in every denominator, and we report full outcome distributions, since a lower lure rate need not imply a higher gold rate.

\vspace{0.1cm}
\noindent\textbf{Models and decoding.} We evaluate six instruction-tuned open-weight models: Qwen3-8B, Qwen3-14B, and Qwen3-32B-AWQ \citep{yang2025qwen3}, Llama-3.1-8B-Instruct \citep{grattafiori2024llama3}, Gemma-3-12B-IT \citep{gemmateam2025gemma3}, and OLMo-2-1124-13B-Instruct \citep{olmo2024olmo2}. We use greedy decoding constrained to a single option letter, which implements the argmax above, and disable the thinking mode of Qwen3. Each model answers all 959 main items under all eight conditions, yielding 46{,}032 responses. Prompts, checkpoints, precision, and other runtime settings appear in Appendix~\ref{app:prompts}, and results on the condition supplement appear in Appendix~\ref{app:condition}.

\subsection{Metrics}
\label{sec:metrics}

Our main outcome is lure selection, that is, how often a misleading context $t \in \{\mathrm S, \mathrm F\}$ leads a model to choose the lure. We measure it in two settings that match the two questions in \S\ref{sec:intro}: with $t$ alone, and with $t$ together with G.

\vspace{0.1cm}
\noindent\textbf{Evaluation subsets.} We focus on answers that a model would otherwise get right. On such items, the model already knows the correct answer, so choosing the lure reflects the influence of the context. For model $m$, $\mathcal B_m$ contains the items it answers correctly without context, and $\mathcal A_m \subseteq \mathcal B_m$ contains those it also answers correctly with G alone, which ensures that the model can use G. Both subsets are fixed from a single response under each of these two conditions, and their sizes are reported in Table~\ref{tab:main-results}.

\vspace{0.1cm}
\noindent\textbf{Misleading context alone.} The lure rate $\mathrm{LR}$ is the fraction of items in $\mathcal B_m$ on which model $m$ chooses the lure when given $t$ alone:
\[
\mathrm{LR}_{m,t} = \frac{1}{|\mathcal B_m|}\sum_{i \in \mathcal B_m} \mathbb{I}\big[\hat{o}_{m,i}(t) = \ell_i\big],
\]
where $\hat{o}_{m,i}(C)$ is the option predicted by model $m$ on item $i$ under context condition $C$ (\S\ref{sec:conditions}), $\ell_i$ is the lure of item $i$, and the indicator $\mathbb{I}[\cdot]$ equals 1 if its argument is true and 0 otherwise.

\vspace{0.1cm}
\noindent\textbf{Misleading context with G.} The residual lure rate $\mathrm{RLR}$ is the fraction of items in $\mathcal A_m$ on which the model still chooses the lure when G is also present, averaged over the two orders:
\[
\begin{aligned}
\mathrm{RLR}_{m,t} &= \frac{1}{2|\mathcal A_m|}\sum_{i \in \mathcal A_m} \Big(\mathbb{I}\big[\hat{o}_{m,i}(\mathrm G, t) = \ell_i\big] +\\
&\qquad \mathbb{I}\big[\hat{o}_{m,i}(t, \mathrm G) = \ell_i\big]\Big).
\end{aligned}
\]
To measure how much G reduces lure selection, we compare RLR with LR computed on the same subset $\mathcal A_m$. We also report the two presentation orders separately, and call their difference, the misleading-first minus G-first rate, the \emph{order effect}.

\subsection{Statistical Inference}
\label{sec:inference}

We compute each metric within each model's subsets and report its average over the six models in $\mathcal M$. Two comparisons are primary because they answer our two questions. They measure how much more F misleads than S, alone and with G:
\[
\Delta_{\mathrm K} = \frac{1}{|\mathcal M|}\sum_{m \in \mathcal M}\big(\mathrm K_{m,\mathrm F} - \mathrm K_{m,\mathrm S}\big),
\]
for $\mathrm K \in \{\mathrm{LR}, \mathrm{RLR}\}$. Other results are descriptive. To estimate uncertainty, we resample clusters of related items (\S\ref{sec:sources}) with replacement, using the same draws for all models and conditions. The two primary comparisons use Bonferroni-adjusted 97.5\% confidence intervals, and all other results use 95\% intervals. As a sensitivity check, we repeat the resampling with 132 coarser groups that also link items sharing documents or distractor sources. Further details appear in Appendices~\ref{app:metrics} and~\ref{app:robustness}.

\section{Results}
\label{sec:results}

\begin{figure}[b]
\centering
\begin{tikzpicture}
\begin{axis}[
  width=\columnwidth,
  height=4.2cm,
  /pgf/number format/assume math mode=true,
  ybar=2.5pt,
  bar width=11pt,
  enlarge x limits=0.2,
  ymin=0, ymax=88,
  ytick={0,20,40,60,80},
  yticklabel={\pgfmathprintnumber{\tick}},
  ylabel={Lure rate (\%)},
  ylabel style={font=\footnotesize, text=inkSecondary},
  symbolic x coords={Alone,G first,Misleading first},
  xtick=data,
  xticklabel style={font=\footnotesize, text=inkPrimary},
  yticklabel style={font=\scriptsize, text=inkSecondary},
  axis lines*=left,
  axis line style={draw=gridGray, line width=0.6pt},
  y axis line style={draw=none},
  major tick length=0pt,
  ymajorgrids,
  grid style={draw=gridGray!55, line width=0.4pt},
  nodes near coords,
  nodes near coords style={font=\scriptsize, text=inkPrimary,
    /pgf/number format/fixed, /pgf/number format/precision=1,
    /pgf/number format/fixed zerofill},
  legend cell align=left,
  legend style={at={(0.98,0.98)}, anchor=north east, draw=none,
    fill=none, font=\footnotesize, text=inkPrimary, row sep=1pt},
  legend image code/.code={\fill[#1] (0cm,-0.09cm) rectangle (0.28cm,0.09cm);},
  clip=false,
]
\addplot[fill=colS, draw=none,
  postaction={pattern={Lines[angle=45, distance=3.2pt, line width=0.55pt]}, pattern color=white!60!colS}]
  coordinates {(Alone,29.23) (G first,0.46) (Misleading first,1.36)};
\addlegendentry{S (true, out-of-scope)}
\addplot[fill=colF, draw=none,
  postaction={pattern={Lines[angle=-45, distance=3.2pt, line width=0.55pt]}, pattern color=white!60!colF}]
  coordinates {(Alone,75.38) (G first,3.91) (Misleading first,38.10)};
\addlegendentry{F (false, in-scope)}
\end{axis}
\end{tikzpicture}
\caption{Lure rates on $\mathcal A_m$ for S and F alone, with G first, and with the misleading context first, using the same items within each model. Rates are equal-weight means across models. Context-alone means differ from the $\mathcal B_m$ rates in \S\ref{sec:main-results} by less than 0.1 percentage points.}
\label{fig:results}
\vspace{-10pt}
\end{figure}
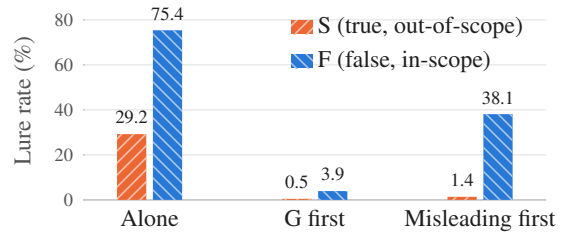

\begin{table*}[t]
\centering
\small
\setlength{\tabcolsep}{6.2pt}
\setlength{\aboverulesep}{0pt}
\setlength{\belowrulesep}{0pt}
\renewcommand{\arraystretch}{1.2}
\begin{tabular}{lcc;{2pt/2pt}ccc>{\columncolor{gray!12}}c;{2pt/2pt}cc>{\columncolor{gray!12}}c}
\toprule
& \multicolumn{2}{c;{2pt/2pt}}{\textbf{Size}} & \multicolumn{4}{c;{2pt/2pt}}{\textbf{Context alone on $\mathcal B_m$}} & \multicolumn{3}{c}{\textbf{With G on $\mathcal A_m$}}\\
\cmidrule(lr){2-3}\cmidrule(lr){4-7}\cmidrule(lr){8-10}
Model & $|\mathcal B_m|$ & $|\mathcal A_m|$ & $\mathrm{Gold}_{\mathrm S}$ & $\mathrm{LR}_{\mathrm S}$ & $\mathrm{LR}_{\mathrm F}$ & $\Delta_{\mathrm{LR}}^{\dagger}$ & $\mathrm{RLR}_{\mathrm S}$ & $\mathrm{RLR}_{\mathrm F}$ & $\Delta_{\mathrm{RLR}}^{\dagger}$\\
\midrule
Qwen3-8B      & 388 & 387 & 55.41 & 18.30 & 63.66 & 45.36 & 0.90 & 14.73 & 13.82\\
Qwen3-14B     & 454 & 454 & 64.10 & 17.40 & 70.93 & 53.52 & 0.66 & 25.99 & 25.33\\
Qwen3-32B-AWQ & 512 & 512 & 71.29 & 24.61 & 85.35 & 60.74 & 0.78 & 27.54 & 26.76\\
Llama-3.1-8B  & 445 & 445 & 59.33 & 26.07 & 77.75 & 51.69 & 0.11 & 18.20 & 18.09\\
Gemma-3-12B   & 494 & 494 & 64.78 & 16.40 & 77.94 & 61.54 & 0.61 & 20.85 & 20.24\\
OLMo-2-13B    & 463 & 459 & 26.13 & 72.79 & 76.67 &  3.89 & 2.40 & 18.74 & 16.34\\
\midrule
\textbf{Mean} &     &     & \textbf{56.84} & \textbf{29.26} & \textbf{75.38} & \textbf{46.12} & \textbf{0.91} & \textbf{21.01} & \textbf{20.10}\\
\textit{CI}   &     &     & {\scriptsize[53.6, 60.1]} & {\scriptsize[26.8, 31.8]} & {\scriptsize[70.8, 79.7]} & {\scriptsize[41.9, 50.0]} & {\scriptsize[0.6, 1.3]} & {\scriptsize[18.8, 23.2]} & {\scriptsize[17.6, 22.6]}\\
\hdashline[1pt/1.5pt]
w/o OLMo      &     &     & 62.98 & 20.55 & 75.12 & 54.57 & 0.61 & 21.46 & 20.85\\
\bottomrule
\end{tabular}
\caption{\textbf{Main results.} $\mathrm{Gold}_{\mathrm S}$ is the gold rate with S alone. LR is the lure rate with S or F alone on $\mathcal B_m$, and RLR is the residual lure rate with G on $\mathcal A_m$, averaged over both orders. $\Delta$ is F minus S in percentage points, and other values are percentages. Means weight models equally, w/o OLMo is the mean without OLMo-2-13B, and values are rounded independently. The CI row gives 95\% cluster-bootstrap intervals, except for the two primary comparisons (shaded, $^\dagger$), which use Bonferroni-adjusted 97.5\% intervals.}
\label{tab:main-results}
\vspace{-10pt}
\end{table*}

We report means over the six models, with intervals computed as described in \S\ref{sec:inference}. Differences are computed before rounding. We first examine misleading context alone (\S\ref{sec:main-results}) and then what changes when gold context (G) is added (\S\ref{sec:resistance-order}).

\subsection{Misleading Context Alone}
\label{sec:main-results}
All results here use each model's own $\mathcal B_m$, the items it answers correctly without any context.

\noindent\textbf{S alone often misleads models.} True but out-of-scope context (S) alone leads models to the lure on 29.26\% of items (95\% CI: [26.79, 31.79]; Table~\ref{tab:main-results}). The corresponding mean gold, abstention, and other-distractor rates are 56.84\%, 12.17\%, and 1.73\%, respectively. The effect extends across models: five show lure rates of 16.40--26.07\% (mean 20.55\%), while OLMo-2-13B reaches 72.79\%. For OLMo-2-13B, the lure rate remains 62.30\% among closed-book-correct items whose S it judges not applicable in the separate scope diagnostic, compared with 6.90--18.75\% for the other models (\S\ref{sec:scope-analysis}).

\vspace{0.1cm}
\noindent\textbf{F misleads more.} With false but in-scope context (F) alone, the lure rate rises to 75.38\%, exceeding S by 46.12 percentage points (97.5\% CI: [41.86, 49.95]). The difference is positive for all six models, ranging from 45.36 to 61.54 points for five and 3.89 points for OLMo-2-13B, misled by S almost as often as by F. Requiring closed-book correctness under three option orders lowers both lure rates, but S still misleads models (\S\ref{sec:heterogeneity}).

\subsection{Misleading Context with G}
\label{sec:resistance-order}

All results in this subsection are computed on each model's $\mathcal A_m$, including the context-alone baselines.

\vspace{0.1cm}
\noindent\textbf{G nearly removes the effect of S but not F.} Adding G reduces the lure rate from 29.23\% to 0.91\% for S, a drop of 28.32 points (95\% CI: [25.92, 30.76]), and from 75.38\% to 21.01\% for F, a drop of 54.37 points ([51.45, 57.13]). Figure~\ref{fig:results} shows these rates separately for each order. Gold selection rises from 56.90\% to 98.82\% with S and from 17.95\% to 77.08\% with F. The residual lure rate of F exceeds that of S by 20.10 points (97.5\% CI: [17.55, 22.56]), and the difference is positive for all six models.

\vspace{0.1cm}
\noindent\textbf{Order matters mainly for F.} With F, the lure rate is 3.91\% when G comes first and 38.10\% when F comes first, a nearly tenfold increase (34.19 points; 95\% CI: [30.20, 38.02]; Figure~\ref{fig:results}). The misleading-first rate is higher for all six models, and the interval stays above zero under dependency-group resampling. With S, the rates are 0.46\% and 1.36\%, an order effect of 0.89 points (95\% CI: [0.33, 1.55]), whose dependency-group interval includes zero (Appendix~\ref{app:robustness}). As a result, the F--S gap in residual lure rate is 3.45 points when G comes first but 36.75 points when the misleading context comes first. Pairing outcomes within each item gives nearly identical rates (Appendix~\ref{app:paired}). This first-context advantage runs counter to a recency effect and suggests a risk for retrieval-augmented settings where passages are presented in rank order: an earlier false passage can remain influential even when a correct one follows. Since both contexts follow the question and options in our prompt, the first is also closer to the question, so our layout cannot separate a primacy effect from proximity to the question.

\section{Analysis}
\label{sec:analysis}
We examine two natural follow-up questions: whether models recognize S as out-of-scope, which could explain why S misleads them (\S\ref{sec:scope-analysis}), and whether an instruction to follow only applicable context, a simple mitigation that requires no training, reduces lure selection (\S\ref{sec:instruction}). We then test robustness across models, sources, and option orders to check that the main findings hold beyond one setup (\S\ref{sec:heterogeneity}). These analyses are exploratory and complement the two primary comparisons in \S\ref{sec:results}, and all rates are equal-weight means over models.

\subsection{Scope Recognition}
\label{sec:scope-analysis}

We sample 200 items, 40 from each source--scope group, independently of model responses. In separate requests, models judge whether each context applies to the question's target scope, regardless of whether it is true, by choosing \emph{Applicable}, \emph{Not applicable}, or \emph{Cannot determine}. The correct label is \emph{Applicable} for G and F and \emph{Not applicable} for S. We then link each judgment to the model's original answers under no context and S alone.

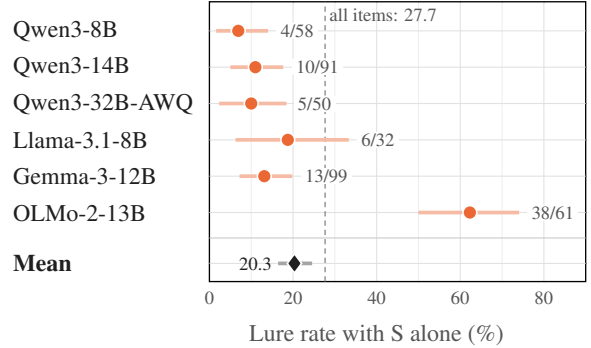
\begin{figure}[!t]
\centering
\pgfplotstableread{
y    rate   em     ep     cnt     px
7    6.90   5.29   7.14   4/58    16.0
6    10.99  5.99   6.66   10/91   19.7
5    10.00  7.67   8.42   5/50    20.1
4    18.75  12.50  14.58  6/32    35.3
3    13.13  5.92   6.66   13/99   21.8
2    62.30  12.30  11.78  38/61   76.0
}\scopeModels
\pgfplotstableread{
y    rate   em     ep
0.6  20.34  3.90   4.22
}\scopeMean
\begin{tikzpicture}
\begin{axis}[
  scale only axis, width=\columnwidth-77.5pt,
  height=3.7cm,
  /pgf/number format/assume math mode=true,
  xmin=0, xmax=90,
  ymin=0.1, ymax=7.8,
  xtick={0,20,40,60,80},
  xticklabel={\pgfmathprintnumber{\tick}},
  xticklabel style={font=\scriptsize, text=inkSecondary},
  xlabel={Lure rate with S alone (\%)},
  xlabel style={font=\footnotesize, text=inkSecondary, at={(ticklabel cs:0.444)}, anchor=near ticklabel},
  ytick={7,6,5,4,3,2,0.6},
  yticklabels={Qwen3-8B,Qwen3-14B,Qwen3-32B-AWQ,Llama-3.1-8B,Gemma-3-12B,OLMo-2-13B,\textbf{Mean}},
  yticklabel style={font=\footnotesize, text=inkPrimary, anchor=west, inner xsep=0pt, xshift=-74pt},
  axis line style={draw=black, line width=0.5pt},
  major tick length=0pt, minor tick length=0pt,
  minor x tick num=1,
  xmajorgrids, xminorgrids,
  major grid style={draw=gridGray!55, line width=0.4pt},
  minor grid style={draw=gridGray!30, line width=0.3pt},
  clip=false,
]
\pgfplotsinvokeforeach{7,6,5,4,3,2,0.6}{
  \draw[draw=gridGray!55, line width=0.4pt] (axis cs:0,#1) -- (axis cs:90,#1);}
\draw[draw=inkSecondary!70, dash pattern=on 2pt off 1.5pt, line width=0.5pt]
  (axis cs:27.67,0.1) -- (axis cs:27.67,7.8);
\node[font=\scriptsize, text=inkSecondary, anchor=west, inner sep=1.5pt]
  at (axis cs:27.67,7.45) {all items: 27.7};
\draw[draw=gridGray, line width=0.4pt] (axis cs:0,1.3) -- (axis cs:90,1.3);
\addplot[only marks, mark=*, mark size=2.6pt,
  mark options={fill=colS, draw=white, line width=0.6pt},
  error bars/.cd, x dir=both, x explicit, error mark=none,
  error bar style={line width=1.4pt, draw=colS!45}]
  table[x=rate, y=y, x error minus=em, x error plus=ep] {\scopeModels};
\addplot[only marks, mark=diamond*, mark size=3.6pt,
  mark options={fill=inkPrimary, draw=white, line width=0.6pt},
  error bars/.cd, x dir=both, x explicit, error mark=none,
  error bar style={line width=1.4pt, draw=inkPrimary!40}]
  table[x=rate, y=y, x error minus=em, x error plus=ep] {\scopeMean};
\node[font=\scriptsize, text=inkPrimary, anchor=east, inner sep=1.5pt]
  at (axis cs:16.0,0.6) {20.3};
\addplot[only marks, mark=none,
  nodes near coords, point meta=explicit symbolic,
  every node near coord/.append style={anchor=west, font=\scriptsize, text=inkSecondary, fill=white, inner xsep=1.5pt, inner ysep=0.5pt}]
  table[x=px, y=y, meta=cnt] {\scopeModels};
\end{axis}
\end{tikzpicture}
\caption{Lure rate with S alone on diagnostic items in $\mathcal B_m$ whose S the same model judges \emph{Not applicable}. Bars show 95\% cluster-bootstrap CIs, and numbers give lure counts and denominators. The diamond and dashed line are the means over models and all items.}
\label{fig:scope-conditional-lure}
\vspace{-10pt}
\end{figure}

\vspace{0.1cm}
\noindent\textbf{Models often miss the scope mismatch.} Models correctly judge G and F as applicable in 100.00\% and 96.00\% of cases, but judge S as not applicable in only 62.42\% (95\% CI: [58.66, 66.09]), even though S states its scope explicitly. \emph{Cannot determine} counts as incorrect.

\vspace{0.1cm}
\noindent\textbf{Recognizing the mismatch does not prevent the lure.} On diagnostic items in $\mathcal B_m$, S alone leads to the lure on 27.67\% of items overall, on 66.06\% of items whose S is judged applicable, and on 20.34\% of items whose S is judged not applicable (95\% CI: [16.44, 24.56]). All six models choose the lure on some items whose S they judge correctly (Figure~\ref{fig:scope-conditional-lure}). Correct judgments go with lower lure rates but do not prevent lure selection. Since judging and answering are separate requests, we cannot tell whether models detect the mismatch while answering. Appendix~\ref{app:scope} gives per-model counts.

\subsection{Context-Faithful Instruction}
\label{sec:instruction}

We re-run all conditions with context using a \emph{context-faithful} (CF) instruction: answer only from applicable context, do not correct it with outside knowledge, and abstain when applicable context is missing or conflicting. Table~\ref{tab:cf-responses} lists the prescribed responses. It prescribes abstention for S alone, because S is not applicable, and for F with G, because the two conflict. For F alone, F is the only applicable context, so CF prescribes the lure. Comparisons use the original evaluation subsets, except that G alone uses all 959 items.

\begin{table}[t]
\centering
\small
\setlength{\tabcolsep}{5.5pt}
\setlength{\aboverulesep}{0pt}
\setlength{\belowrulesep}{0pt}
\renewcommand{\arraystretch}{1.4}
\begin{tabular}{llcc>{\columncolor{gray!12}}c}
\toprule
\textbf{Condition} & \textbf{Prescribed} & \textbf{Original} & \textbf{CF} & $\boldsymbol{\Delta}$\\
\midrule
S alone  & \textit{Abstain} & 12.17 & 77.85 & $+$65.68\\
F alone  & \textit{Lure}    & 75.38 & 69.43 & $-$5.95\\
G alone  & \textit{Gold}    & 98.68 & 91.26 & $-$7.42\\
\hdashline[1pt/1.5pt]
S with G & \textit{Gold}    & 98.82 & 93.84 & $-$4.98\\
F with G & \textit{Abstain} &  1.74 & 44.48 & $+$42.74\\
\bottomrule
\end{tabular}
\caption{Rate (\%) of the response CF prescribes for each condition under both prompts. $\Delta$ is CF minus original in percentage points. Conditions with G average both orders. Gold and lure keep their factual meaning, so the prescribed F-alone response is incorrect.}
\label{tab:cf-responses}
\vspace{-10pt}
\end{table}

\vspace{0.1cm}
\noindent\textbf{CF reduces lure selection mainly through abstention.} With S alone, CF lowers the lure rate from 29.26\% to 15.06\%, but abstention rises from 12.17\% to 77.85\% and gold selection falls from 56.84\% to 6.99\%. Abstention is the prescribed response here, yet models still choose the lure on 15.06\% of items. With F and G, abstention also rises, from 1.74\% to 44.48\%, but stays below half.

\vspace{0.1cm}
\noindent\textbf{CF lowers accuracy when G is available.} Gold selection falls from 98.68\% to 91.26\% with G alone and from 98.82\% to 93.84\% with S and G, although gold is the prescribed response in both conditions. The drop with S and G persists, from 98.92\% to 94.82\%, on items answered correctly with G alone under both prompts. CF changes several requirements at once, so these results do not isolate the effect of any single clause. The full prompt and per-model results appear in Appendix~\ref{app:faithful}.

\subsection{Robustness}
\label{sec:heterogeneity}

\noindent\textbf{Models and sources.} Excluding OLMo-2-13B lowers the lure rate with S alone from 29.26\% to 20.55\%, and raises the F--S differences to 54.57 points for LR and 20.85 points for RLR. Across source--scope groups, the lure rate with S alone ranges from 7.59\% for SituatedQA location to 73.62\% for AmbigDocs entity. Since sources and scopes are not fully crossed, these group differences may also reflect source content and wording. Further checks appear in Appendix~\ref{app:robustness}.

\vspace{0.1cm}
\noindent\textbf{Option order.} On the 200-item subset, we collect new responses under the original option order and two cyclic rotations of A--D, with E fixed. Under the original order, the new responses match the main evaluation exactly. Requiring correctness without context under all three orders keeps 45--68 items per model, or 62.22\% of those answered correctly under the original order. This stricter subset lowers both lure rates, but the F--S difference persists (Table~\ref{tab:option-robustness}). Averaged over the three orders, it is 40.38 points (95\% CI: [33.69, 47.42]; dependency-group CI: [29.63, 49.94]). Excluding OLMo-2-13B, lure rates are 9.50\% for S and 56.29\% for F. This check excludes conditions with G.

\begin{table}[t]
\centering
\small
\setlength{\tabcolsep}{6.5pt}
\setlength{\aboverulesep}{0pt}
\setlength{\belowrulesep}{0pt}
\renewcommand{\arraystretch}{1.4}
\begin{tabular}{llcc>{\columncolor{gray!12}}c}
\toprule
\multicolumn{2}{c}{\textbf{Order used for}} & \multicolumn{3}{c}{\textbf{Lure rate (\%)}}\\
\cmidrule(lr){1-2}\cmidrule(lr){3-5}
\textbf{Selection} & \textbf{Measurement} & \textbf{S} & \textbf{F} & \textbf{F$-$S}\\
\midrule
Original & Original & 27.67 & 71.10 & 43.43\\
\hdashline[1pt/1.5pt]
\multirow{2}{*}{All three} & Original      & 18.39 & 55.27 & 36.88\\
                           & Mean of three & 17.55 & 57.93 & 40.38\\
\bottomrule
\end{tabular}
\caption{Option-order robustness on the 200-item subset. Items are selected if answered correctly without context under the original order or under all three orders, and lure rates are measured under the original order or averaged over the three. S and F use the same items within each model, and F$-$S is in percentage points.}
\label{tab:option-robustness}
\vspace{-10pt}
\end{table}

\section{Conclusion}
\label{sec:conclusion}

We contributed \textsc{MisleadQA}, a dataset that pairs true but out-of-scope context (S) with false but in-scope context (F) supporting the same wrong answer, with the question, options, and gold context (G) fixed. Across six open-weight LLMs, S alone often misleads models on questions they otherwise answer correctly, and F misleads them more. Adding G nearly removes the effect of S but not that of F, especially when F comes first. Models choose the lure under S even when they separately judge S as out-of-scope, and a context-faithful instruction reduces misled answers mainly through abstention while lowering accuracy with G. These results suggest that evaluations should check whether context is both true and applicable to the question, and report correct, misled, and abstention rates together, since a single rate gives an incomplete picture.

\clearpage
\section*{Limitations}

\textsc{MisleadQA} enables a controlled comparison of out-of-scope and false context on the same questions. Our findings show that checking whether context is true is not enough, and that lure rates alone can hide lost correct answers. These conclusions are drawn under several limitations.

\vspace{0.1cm}
\noindent\textbf{Models.} We evaluate six open-weight models from four families, with 8B to 32B parameters, and disable the thinking mode of Qwen3. All six models are misled more by F than by S, both alone and with G, and all are misled more when F comes before G. Larger proprietary models and models that reason before answering remain to be tested.

\noindent\textbf{Task format.} Each context is a single claim that states its scope explicitly, and models choose among five options with constrained decoding. This format keeps S and F minimally different and makes every answer unambiguous to score. Real retrieval-augmented systems often use long, multi-passage documents and free-form answers, where scope cues may be implicit. Since explicit cues should make mismatches easier to detect, our setting may understate how often out-of-scope context misleads models in practice.

\vspace{0.1cm}
\noindent\textbf{Dataset design.} S and F differ in both truth and scope, so their comparison reflects the joint effect of the two factors. Separating them would require context that is both false and out-of-scope. In addition, sources and scope dimensions are only partly crossed, since all entity mismatches in the main split come from AmbigDocs.

\section*{Ethics Statement}

\noindent\textbf{Data sources and release.} \textsc{MisleadQA} is built from public resources: state minimum-wage records from the U.S.\ Department of Labor, which are in the public domain and accessed through FRED, and the SituatedQA, AmbigDocs (Apache 2.0), and CondAmbigQA (MIT) datasets. SituatedQA specifies no license, so we use it for research with citation. The data cover public facts about states, events, and named entities and contain no private personal information. Because some contexts come from Wikipedia, we release the dataset under CC BY-SA 4.0 and the code under MIT.

\vspace{0.1cm}
\noindent\textbf{False context.} Each item includes a false context (F), which is explicitly labeled as false in the released data. F is intended only for evaluating how models respond to misinformation, and the dataset should not be used as a source of facts or as training data for these claims.

\vspace{0.1cm}
\noindent\textbf{Human annotation.} Six in-house annotators audited a sample of 150 items. Before starting, each annotator gave informed consent by confirming a consent statement. We collected no personal information from the annotators. 

\vspace{0.1cm}
\noindent\textbf{Use of AI assistants.} GPT-6 Astra helped word some restated contexts, and GPT-6 Astra and Claude Opus 5.5 screened items for errors, with every flagged item reviewed by a human (\S\ref{sec:construction} and \S\ref{sec:qc}). We also used AI assistants to polish author-written text and to help with coding. The authors reviewed all suggested changes and take full responsibility for the paper and code.

\bibliography{custom, anthology-1, anthology-2}

\clearpage
\appendix
\section{Dataset Construction}
\label{app:dataset-construction}

This section describes how we construct the 1,100-item candidate pool. Appendix~\ref{app:construction} gives the construction steps and the criteria each item must meet, using one FRED item as a running example from fact selection to the complete item. Appendix~\ref{app:options-pool} describes the answer options and the resulting candidate pool, Appendix~\ref{app:examples} shows one example from each source, and Appendix~\ref{app:provenance} describes the source records and the clusters used for resampling. All candidates are then screened against these criteria, and Appendix~\ref{app:quality-assurance} reports the screening results, revisions, and release selection.

\vspace{0.1cm}
\noindent\textbf{Sources.} FRED provides archived annual observations of the basic state minimum wage for 44 jurisdictions, namely 43 states and the District of Columbia \citep{fred_minimum_wage}. The series originate from the U.S.\ Department of Labor. G and S facts use observations from 1990 to 2024, and questions ask for the recorded value in US dollars per hour on a given date, excluding worker coverage, exemptions, and changes within a year. SituatedQA \citep{zhang-choi-2021-situatedqa} provides questions whose answers depend on a specific time or location. Because its original answer labels alone are not sufficient, we verify each answer against archived passages or structured records. AmbigDocs \citep{lee2024ambigdocs} provides Wikipedia excerpts about distinct entities that share a name, and CondAmbigQA \citep{li-etal-2025-condambigqa} supplies the condition candidates, which are kept separate from the main candidates.

\subsection{Item Construction}
\label{app:construction}

All sources follow the same construction guideline. As in \S\ref{sec:construction}, each item starts from a fact pair and is built in three steps: writing the question, creating the contexts, and choosing the answer options. We then select candidates across items to limit over-representation. We illustrate each step with a FRED time item about Nebraska, shown in full in Table~\ref{tab:example-fred}.

\vspace{0.1cm}
\noindent\textbf{Fact pairs.} Each item starts from two source-supported facts about the same attribute at two scopes. The scopes differ in exactly one designated dimension, and all other qualifiers stay fixed. At each scope, the answer must be unique at the precision the question asks for. The fact at the target scope gives the gold answer, and the fact at the alternative scope gives the lure, which must be wrong at the target scope and distinct from the gold answer after normalizing aliases, formats, and units. We fix both scopes before writing the question, and for FRED time pairs the two dates are at most eight years apart. In the running example, the Nebraska rate on January 1, 1997 gives the gold answer (\$4.25), and the rate on January 1, 1998 gives the lure (\$5.15), so only the date changes.

\vspace{0.1cm}
\noindent\textbf{Questions and contexts.} Each question names the attribute and the target scope without revealing the gold answer or the lure. G states the gold answer at the target scope, and S states the lure at an explicitly named alternative scope. If S gives a scope range, the range must exclude the target scope. We build G and S by verbalizing records with fixed templates, extracting contiguous source text, or restating source content without changing its meaning, and we record the method and keep the supporting source material for each context. For SituatedQA, we also record which scope and answer annotations come from the original dataset and which are added or corrected from archived sources. F is derived from G by replacing every registered expression of the gold answer, including aliases, with the lure and leaving all other text unchanged. The edit must leave no gold mention, keep the grammar intact, and add no contradiction beyond the false answer. We record each substitution and do not treat F as source-supported. In the running example, replacing \$4.25 with \$5.15 in G yields F, which keeps the 1997 date.

\begin{table}[t]
\centering
\small
\setlength{\tabcolsep}{4pt}
\setlength{\aboverulesep}{0pt}
\setlength{\belowrulesep}{0pt}
\renewcommand{\arraystretch}{1.25}
\begin{tabular}{>{\columncolor{gray!12}\bfseries}p{0.18\columnwidth}>{\raggedright\arraybackslash}p{0.75\columnwidth}@{}}
\toprule
Question & What was the recorded basic state minimum wage in Nebraska on January~1, 1997? Give the rate in US dollars per hour.\\
\hdashline[1pt/1.5pt]
Options & A.~\$5.15~\textit{(lure)}\quad B.~\$13.50\quad C.~\$4.25~\textit{(gold)}\quad D.~\$5.85\quad E.~Insufficient~information\\
\midrule
G & On January~1, 1997, the recorded basic state minimum wage in Nebraska was \$4.25 per hour.\\
\hdashline[1pt/1.5pt]
F & On January~1, 1997, the recorded basic state minimum wage in Nebraska was \textbf{\$5.15} per hour.\\
\hdashline[1pt/1.5pt]
S & On January~1, \textbf{1998}, the recorded basic state minimum wage in Nebraska was \textbf{\$5.15} per hour.\\
\bottomrule
\end{tabular}
\caption{FRED time example. G and S verbalize archived observations, and F replaces only the rate in G. Bold marks where F and S differ from G.}
\label{tab:example-fred}
\vspace{-10pt}
\end{table}

\vspace{0.1cm}
\noindent\textbf{Context comparability.}\label{app:fs-similarity} We check that S is not easier to match to the question than G or F, so that its effect does not come from surface overlap. For length, we compare the longest and shortest context in each G/S/F triple by word-count ratio and difference. For SituatedQA, contexts have 7--115 words, with a ratio of at most 1.35 and a difference of at most 25 words. For overlap, we compute Jaccard similarity and question-token recall between the question and each context on normalized token sets without stop words. Jaccard similarity is the number of shared tokens divided by the number of distinct tokens in both sets, and recall is the share of distinct question tokens that appear in the context. On both metrics, S must score no higher than G and F. These checks limit surface differences but do not guarantee equal difficulty.

\vspace{0.1cm}
\noindent\textbf{Distractors.} Each item also has two distractors, each the source-supported value of the same attribute at another scope and wrong at the target scope. The four substantive options must stay distinct after normalization and be comparable in answer type and precision, and no distractor may appear in or be supported by the question or contexts. For FRED, time items take distractors from other dates in the same jurisdiction, which may fall outside 1990--2024 and are not bound by the eight-year limit, and location items take them from other jurisdictions on the target date. For SituatedQA and AmbigDocs, we prefer distractors from the same upstream question and use other questions only when sources confirm the same relation, while CondAmbigQA distractors always come from the same upstream question. Items with fewer than two valid distractors are deferred. In the running example, the 2008 and 2025 observations give the distractors \$5.85 and \$13.50.

\vspace{0.1cm}
\noindent\textbf{Candidate selection.} We cap the number of candidates per source to avoid over-representing any jurisdiction or question. For FRED, we select at most six items per target jurisdiction in each scope dimension. For SituatedQA, we select at most two items per upstream question and twenty per question type in each scope dimension, after merging equivalent question-type labels. For AmbigDocs and CondAmbigQA, we select at most one and three items per upstream question, respectively. We also exclude repeated target facts, including rewordings of the same question, and reversed versions of an existing fact pair. Model responses and observed lure effects play no role in candidate selection.

\subsection{Answer Options and Candidate Pool}
\label{app:options-pool}

\noindent\textbf{Answer options.} Each candidate has five options. Options A--D hold the gold answer, the lure, and the two additional distractors in distinct positions, and option E is always \emph{Insufficient information}, used for abstention. The positions set during construction are provisional, because gold and lure positions are balanced after release selection (Appendix~\ref{app:positions}). The example tables show the positions recorded at construction. In the Nebraska candidate (Table~\ref{tab:example-fred}), the gold answer is option C, the lure is option A, and the additional distractors are options B and D.

\vspace{0.1cm}
\noindent\textbf{Candidate pool.} Construction yields 1,100 candidates: 200 each for FRED time, FRED location, SituatedQA time, SituatedQA location, and AmbigDocs entity, plus 100 CondAmbigQA condition candidates. We save all candidates and their construction metadata in a snapshot with a version number and file hashes, and record later changes separately. This snapshot is the input to screening and adjudication (Appendix~\ref{app:quality-assurance}).

\subsection{Construction Examples}
\label{app:examples}

Tables~\ref{tab:example-fred}--\ref{tab:example-cond} show one candidate from each source, as recorded in the candidate snapshot before screening and revision. Released items may differ in wording or option order. Table~\ref{tab:example-fred} shows the complete Nebraska item from Appendix~\ref{app:construction}, and the other three tables show how construction differs across sources.

\vspace{0.1cm}
\noindent\textbf{SituatedQA.} In Table~\ref{tab:example-situatedqa}, the question asks who was prime minister of India on February~10, 2012. G and S restate source-supported facts and make the office, date, and answer explicit. G refers to the target date, while S states the prime minister in 2002. F keeps the target date and replaces the answer in G with the lure from S.

\begin{table}[t]
\centering
\small
\setlength{\tabcolsep}{4pt}
\setlength{\aboverulesep}{0pt}
\setlength{\belowrulesep}{0pt}
\renewcommand{\arraystretch}{1.25}
\begin{tabular}{>{\columncolor{gray!12}\bfseries}p{0.18\columnwidth}>{\raggedright\arraybackslash}p{0.75\columnwidth}@{}}
\toprule
Question & Who served as prime minister of India on February~10,~2012?\\
\hdashline[1pt/1.5pt]
Options & A.~Jawaharlal~Nehru\quad B.~Narendra~Damodardas~Modi\quad C.~Atal~Bihari~Vajpayee~\textit{(lure)}\quad D.~Manmohan~Singh~\textit{(gold)}\quad E.~Insufficient~information\\
\midrule
G & Manmohan Singh served as prime minister of India on February~10,~2012.\\
\hdashline[1pt/1.5pt]
F & \textbf{Atal Bihari Vajpayee} served as prime minister of India on February~10,~2012.\\
\hdashline[1pt/1.5pt]
S & \textbf{Atal Bihari Vajpayee} served as prime minister of India \textbf{in 2002}.\\
\bottomrule
\end{tabular}
\caption{SituatedQA time example. S states the prime minister in 2002, and F assigns that person to the target date. Bold marks where F and S differ from G.}
\label{tab:example-situatedqa}
\vspace{-10pt}
\end{table}

\vspace{0.1cm}
\noindent\textbf{AmbigDocs.} In Table~\ref{tab:example-ambigdocs}, two townships in Ohio share the name Adams Township. G describes the one in Clinton County, and S describes the one in Coshocton County. Both report the 2010 census population, and the county name serves as the entity identifier, so the two scopes differ in entity. F keeps the target township and replaces its population with the lure from S.

\begin{table}[t]
\centering
\small
\setlength{\tabcolsep}{4pt}
\setlength{\aboverulesep}{0pt}
\setlength{\belowrulesep}{0pt}
\renewcommand{\arraystretch}{1.25}
\begin{tabular}{>{\columncolor{gray!12}\bfseries}p{0.18\columnwidth}>{\raggedright\arraybackslash}p{0.75\columnwidth}@{}}
\toprule
Question & What was the population of Adams Township, Ohio in Clinton County at the 2010 census?\\
\hdashline[1pt/1.5pt]
Options & A.~2,036\quad B.~790~\textit{(lure)}\quad C.~1,110\quad D.~2,091~\textit{(gold)}\quad E.~Insufficient~information\\
\midrule
G & Adams Township is one of the thirteen townships of Clinton County, Ohio, United States. As of the 2010 census the population was 2,091.\\
\hdashline[1pt/1.5pt]
F & Adams Township is one of the thirteen townships of Clinton County, Ohio, United States. As of the 2010 census the population was \textbf{790}.\\
\hdashline[1pt/1.5pt]
S & Adams Township is one of the \textbf{twenty-two} townships of \textbf{Coshocton County}, Ohio, United States. As of the 2010 census the population was \textbf{790}.\\
\bottomrule
\end{tabular}
\caption{AmbigDocs entity example. S describes a township with the same name in another county, and F assigns its population to the target township. Bold marks where F and S differ from G.}
\label{tab:example-ambigdocs}
\end{table}

\vspace{0.1cm}
\noindent\textbf{CondAmbigQA.} In Table~\ref{tab:example-cond}, the two scopes differ in release format for the same film. The question, G, and F concern the digital download release, while S states the Blu-ray and DVD release. F keeps the release format of G and replaces the date with the lure from S. Condition candidates are kept separate from the main candidates and form the condition supplement.

\begin{table}[t]
\centering
\small
\setlength{\tabcolsep}{4pt}
\setlength{\aboverulesep}{0pt}
\setlength{\belowrulesep}{0pt}
\renewcommand{\arraystretch}{1.25}
\begin{tabular}{>{\columncolor{gray!12}\bfseries}p{0.18\columnwidth}>{\raggedright\arraybackslash}p{0.75\columnwidth}@{}}
\toprule
Question & When was Star Wars: The Last Jedi released through digital download?\\
\hdashline[1pt/1.5pt]
Options & A.~March~27,~2018~\textit{(lure)}\quad B.~December~26,~2019\quad C.~December~15,~2017\quad D.~March~13,~2018~\textit{(gold)}\quad E.~Insufficient~information\\
\midrule
G & Star Wars: The Last Jedi was released through digital download on March~13,~2018.\\
\hdashline[1pt/1.5pt]
F & Star Wars: The Last Jedi was released through digital download on \textbf{March~27,~2018}.\\
\hdashline[1pt/1.5pt]
S & Star Wars: The Last Jedi was released \textbf{on Blu-ray and DVD} on \textbf{March~27,~2018}.\\
\bottomrule
\end{tabular}
\caption{CondAmbigQA condition example. S describes the Blu-ray and DVD release of the same film, and F assigns its date to the digital download. Bold marks where F and S differ from G.}
\label{tab:example-cond}
\vspace{-10pt}
\end{table}

\subsection{Source Provenance and Clusters}
\label{app:provenance}

\noindent\textbf{Provenance records.} Each candidate record keeps source identifiers and locators for G, S, and the two additional distractors, together with archived-file hashes and source revision information when available. Supporting source material is stored separately from the rendered contexts, so verbalized records and restatements can be traced back to their sources. For each edit from G to F, we record the edited spans and replacement strings and link them to the version of G they modify.

\vspace{0.1cm}
\noindent\textbf{Source clusters.}\label{app:sources} A source cluster groups items that share a target jurisdiction for FRED or an upstream question for the other sources, so FRED time and location items about the same jurisdiction fall in the same cluster. The 959 main items form 489 source clusters: 44 from FRED, 253 from SituatedQA, and 192 from AmbigDocs. These clusters are the resampling units for the main confidence intervals (Appendix~\ref{app:metrics}).

\vspace{0.1cm}
\noindent\textbf{Dependency groups.} For sensitivity analyses, we also use coarser dependency groups, which merge source clusters whose items share source documents or distractor sources. The 489 source clusters form 132 dependency groups; each cluster lies entirely within one group, and all 400 FRED items form a single group. The released membership index records both groupings, identifying source clusters by source and \texttt{cluster\_id} and dependency groups by \texttt{analysis\_group\_id}. Membership is fixed across models and evaluation conditions, and Appendix~\ref{app:robustness} reports intervals under dependency-group resampling. The condition supplement is analyzed separately and has 20 source clusters and 20 dependency groups.

\section{Quality Assurance and Human Audit}
\label{app:quality-assurance}

Quality assurance has two stages. Screening checks all 1,100 candidates and decides which items enter the release (Appendix~\ref{app:screening}). The human audit then checks a sample of released main items against the intended design of G, S, and F, and does not change the release (Appendix~\ref{app:audit}).

\subsection{Screening and Release Selection}
\label{app:screening}

\noindent\textbf{Automated checks.} Scripts check item structure, answer distinctness after normalization, answer isolation, distractor leakage, source-reference fields, context hashes, and release checksums. Answer isolation requires that neither the gold answer nor the lure appears in the question. The scripts also replay the registered answer substitutions, which must reproduce F exactly from G and leave all text outside the edited spans unchanged. No candidate in the 1,100-item pool fails these checks.

\vspace{0.1cm}
\noindent\textbf{Screening models and criteria.} Two models outside the evaluation panel screen all candidates: GPT-6 Astra in a fresh session without construction history, and Claude Opus 5.5, which was not used in construction. Each model judges every criterion below as \texttt{pass}, \texttt{fail}, or \texttt{uncertain}, with a short reason. A failure marks a definite violation, and an uncertain judgment marks missing evidence or an unclear interpretation. The models only report findings, and a human makes every retention decision.

\begin{enumerate}[leftmargin=*,itemsep=2pt,topsep=3pt]
\item \textbf{Source support.} G, S, and both distractors are supported by their sources for the subject, queried property, scope, and answer. Restatements, aliases, and format conversions keep the supported meaning and precision.
\item \textbf{Explicit scopes.} The question states the target scope without revealing the gold answer or the lure. G and F state the target scope, and S states an alternative scope under which the lure holds.
\item \textbf{Single-dimension change.} S differs from G in exactly one designated scope dimension and keeps the queried property and all other qualifiers.
\item \textbf{Answer uniqueness.} Each scope supports a unique answer at the requested precision, and the gold answer is the only correct option among A--D for the target scope. Options A--D differ in meaning and stay distinct after normalization.
\item \textbf{F edits and fluency.} F equals G except for the registered gold-to-lure substitutions. It keeps the target qualifiers, is grammatical, and contains no remaining gold expression or other unintended contradiction.
\item \textbf{Distractor validity.} Both distractors have documented support, follow the distractor rules in Appendix~\ref{app:construction}, and are wrong for the target scope. They match the gold answer in type and precision and do not appear in the question or contexts.
\end{enumerate}

\noindent\textbf{Screening flags.} An item is flagged when a model marks at least one criterion as \texttt{fail}, and each item counts once regardless of how many criteria fail. GPT-6 Astra flags 118 items and Claude Opus 5.5 flags 239, and 91 items are flagged by both, giving $118+239-91=266$ flagged items. Table~\ref{tab:screening-coverage} breaks these counts down by source. No FRED item is flagged, and most flags fall on SituatedQA and CondAmbigQA.

\begin{table}[t]
\centering
\small
\setlength{\tabcolsep}{7pt}
\setlength{\aboverulesep}{0pt}
\setlength{\belowrulesep}{0pt}
\renewcommand{\arraystretch}{1.2}
\begin{tabular}{lc;{2pt/2pt}cc>{\columncolor{gray!12}}c}
\toprule
& & \multicolumn{3}{c}{\textbf{Flagged items}}\\
\cmidrule(lr){3-5}
\textbf{Source} & \textbf{Items} & \textbf{Astra} & \textbf{Opus} & \textbf{Either}\\
\midrule
FRED        & 400   & 0   & 0   & 0\\
SituatedQA  & 400   & 38  & 162 & 168\\
AmbigDocs   & 200   & 8   & 15  & 23\\
CondAmbigQA & 100   & 72  & 62  & 75\\
\hdashline[1pt/1.5pt]
Total       & 1,100 & 118 & 239 & 266\\
\bottomrule
\end{tabular}
\caption{Screening flags by source. Both models screen all candidates, and an item is flagged if a model marks at least one criterion as \texttt{fail}. Astra is GPT-6 Astra, Opus is Claude Opus 5.5, and Either counts items flagged by at least one model.}
\label{tab:screening-coverage}
\vspace{-10pt}
\end{table}

\noindent\textbf{Human adjudication.}\label{app:adjudication} A human reviewer inspects all 266 flagged items and records a decision and reason for each. The reviewer retains 155 items whose flagged problems are not confirmed, excludes 64, marks 44 for revision, and marks 3 for further verification. The 64 exclusions comprise 62 items with problems in the condition pair, 1 item with overlapping options, and 1 item that changes both release format and geographic market.

\vspace{0.1cm}
\noindent\textbf{Release membership.}\label{app:release-membership} A later review withholds 50 further candidates: the 44 revision cases, the 3 verification cases, and 3 retained condition items that were reopened. This review covers items that were already screened and adds no new screening coverage. Reconciling item IDs against the release identifies 2 more candidates that were not released, leaving $1{,}100-64-50-2=984$ released items: 959 in the main split and 25 in the condition supplement. Overall, the main split loses 41 of its 1,000 candidates, and the supplement loses 75 of its 100.

\vspace{0.1cm}
\noindent\textbf{Final option assignment.}\label{app:positions} After the item set is fixed, we balance gold and lure positions within each source--scope group of the main split. For gold and lure separately, counts across A--D differ by at most one, and counts across the 12 ordered gold--lure position pairs also differ by at most one. Option E is always \emph{Insufficient information}.

\vspace{0.1cm}
\noindent\textbf{Frozen release.}\label{app:final-release} All model evaluations use the same frozen \textsc{MisleadQA} v1.0.0 release, which stores the final option assignments and contains 959 main items and 25 condition-supplement items. The two splits are analyzed separately.

\subsection{Post-freeze Human Audit}
\label{app:audit}

\noindent\textbf{Sample.} The audit uses a stratified sample of 150 released main items (Table~\ref{tab:audit-strata}): 15 from each FRED scope, 35 from each SituatedQA scope, and 50 AmbigDocs entity items. The condition supplement is not audited. The sample was fixed before annotation began, with items selected within each stratum by SHA-256 ranking under seed \texttt{20261007}. Model responses and error labels are not used in sampling. Separate hash domains set the annotator pairs and each annotator's display order, and the sampling code and manifest record the source, sample, and assignment hashes.

\begin{table}[t]
\centering
\small
\setlength{\tabcolsep}{5pt}
\setlength{\aboverulesep}{0pt}
\setlength{\belowrulesep}{0pt}
\renewcommand{\arraystretch}{1.2}
\begin{tabular}{lcc>{\columncolor{gray!12}}cc}
\toprule
\textbf{Stratum} & \textbf{Main} & \textbf{Sampled} & \textbf{Pass} & \textbf{Issue}\\
\midrule
FRED time           & 200 & 15  & 15  & 0\\
FRED location       & 200 & 15  & 15  & 0\\
\hdashline[1pt/1.5pt]
SituatedQA time     & 183 & 35  & 35  & 0\\
SituatedQA location & 184 & 35  & 35  & 0\\
\hdashline[1pt/1.5pt]
AmbigDocs entity    & 192 & 50  & 49  & 1\\
\midrule
Total               & 959 & 150 & 149 & 1\\
\bottomrule
\end{tabular}
\caption{Human-audit sample and outcomes by stratum. Main counts released main items, Pass counts items for which both annotators answer Yes to Q1--Q3 before adjudication, and Issue counts confirmed issues.}
\label{tab:audit-strata}
\vspace{-10pt}
\end{table}

\vspace{0.1cm}
\noindent\textbf{Annotators and assignment.} Six in-house annotators recruited within our research group carry out the audit. Each annotator reads the task description and gives informed consent before starting, and no personal information is collected. None of the annotators took part in item construction. Each item is assigned to two annotators, each annotator receives 50 items, and every pair of annotators shares exactly 10 items, giving 300 assignments. Annotators work independently, without discussing items or consulting external materials. For each item, they see the question, all options, the gold answer and the lure, the target and alternative scopes, and G, S, and F. Source facts are assumed correct, so the audit checks whether the texts follow the intended construction without re-verifying factual labels.

\vspace{0.1cm}
\noindent\textbf{Audit criteria.} Annotators answer three Yes/No questions for each item:

\begin{enumerate}[leftmargin=*,itemsep=2pt,topsep=3pt]
\item[] \textbf{Q1: G support.} Does G clearly support the gold answer for the target scope?
\item[] \textbf{Q2: S support and scope.} Does S clearly support the lure for the same queried property, under a scope that differs from G in exactly one element? Missing information or vague wording alone does not count as a scope change.
\item[] \textbf{Q3: F support and edits.} Does F clearly support the lure for the same scope as G, and match G except that the gold-answer expressions are replaced by the lure? Annotators check for extra edits, remaining gold expressions, and broken grammar.
\end{enumerate}

\noindent An annotator answers Yes only when the whole criterion is met. Every No comes with a note and an item-level label, \emph{Clear issue} or \emph{Unsure}, and an item with both a clear issue and an uncertain check is labeled \emph{Unsure} with the affected checks named. A review passes when Q1--Q3 are all Yes, and an item passes when both of its reviews pass. These criteria are narrower than the six screening criteria, since they do not re-verify source facts or fully assess the distractors.

\vspace{0.1cm}
\noindent\textbf{Completion and agreement.} All six annotators completed their assignments, giving 300 reviews and 900 Yes/No judgments with none missing and no \emph{Unsure} labels. Submission receipts, assignment coverage, and displayed inputs were checked against the fixed sample and assignments. Both annotators answer Yes to all three questions for 149 of the 150 items (99.33\%). For the remaining item, one annotator answers No to Q1 and Q3 and labels it \emph{Clear issue}, while the other answers Yes to all three. Table~\ref{tab:audit-checks} reports the judgments before adjudication. Agreement is also 149/150 (99.33\%) at the item level. We report raw agreement because nearly all judgments are Yes, which makes chance-corrected coefficients such as Krippendorff's $\alpha$ unstable.

\begin{table}[h]
\centering
\small
\setlength{\tabcolsep}{9pt}
\setlength{\aboverulesep}{0pt}
\setlength{\belowrulesep}{0pt}
\renewcommand{\arraystretch}{1.2}
\begin{tabular}{lc>{\columncolor{gray!12}}c}
\toprule
\textbf{Criterion} & \textbf{Yes} & \textbf{Agreement (\%)}\\
\midrule
Q1: G support                  & 299 & 99.33\\
Q2: S support and scope        & 300 & 100.00\\
Q3: F support and edits        & 299 & 99.33\\
\bottomrule
\end{tabular}
\caption{Human-audit judgments before adjudication. Each criterion has 300 judgments on 150 items. Yes counts Yes judgments, and agreement is the share of items where both annotators give the same answer.}
\label{tab:audit-checks}
\vspace{-10pt}
\end{table}

\vspace{0.1cm}
\noindent\textbf{Adjudication.} A third annotator adjudicates the disputed item and confirms a construction issue.\footnote{\texttt{ambigdocs\_entity\_f96e8711d9ee121a\_v8}} The issue concerns whether the person named in the question, Jamie Elliott, is the same as the person named in G and F, James Harry Elliott. Both No judgments concern this item, so the confirmed issue rate is $1/150=0.67\%$. Adjudication sets the final outcome but does not change the original judgments or the agreement statistics. The audit does not change the frozen release, so this item and the recorded model responses remain unchanged.

\section{Experimental Details}
\label{app:evaluation-protocol}

This section gives the prompts, models, and decoding settings used in all experiments (Appendix~\ref{app:prompts}) and the cluster-bootstrap procedure used to compute all confidence intervals (Appendix~\ref{app:metrics}).

\subsection{Prompts, Models, and Decoding}
\label{app:prompts}

\noindent\textbf{Answering prompt.} Every request is a single user turn built from the template in Figure~\ref{fig:prompt}. Each option appears on its own line as the letter, a period, a space, and the option text. Contexts follow the options in the order of the condition, each preceded by a blank line, and are omitted in the no-context condition. The prompt contains no G, F, or S labels, answer keys, or construction metadata, and it gives no instruction on how to use the contexts. Each condition is sent as a separate request without shared history.

\begin{figure}[t]
\begin{lstlisting}[style=prompt]
Answer the following multiple-choice question. Return only A, B, C, D, or E.

Question:
(*\promptfield{question}*)

Options:
(*\promptfield{options}*)

(*\promptfield{context 1}*)

(*\promptfield{context 2}*)
\end{lstlisting}
\caption{Answering prompt used in all eight conditions. Blue fields are filled for each request. Without context, the prompt ends after the options, and with one context, only the first context field appears.}
\label{fig:prompt}
\vspace{-10pt}
\end{figure}

\vspace{0.1cm}
\noindent\textbf{Analysis prompts.} The scope analysis in \S\ref{sec:scope-analysis} uses the prompt in Figure~\ref{fig:prompt-analysis}(a). Each request contains the question, its A--E options, and one context, and G, S, and F are judged in separate requests without their labels or the answer key. The context-faithful (CF) analysis in \S\ref{sec:instruction} uses the prompt in Figure~\ref{fig:prompt-analysis}(b), which adds an instruction and fills the other fields as in Figure~\ref{fig:prompt}. It is used in the seven conditions with context, and the no-context condition keeps the answering prompt.

\vspace{0.1cm}
\noindent\textbf{Chat templates.} We wrap every prompt in the checkpoint's official chat template and append the assistant generation prefix. For Qwen3, we set \texttt{enable\_thinking=False} to disable thinking mode. The date in the Llama template is fixed to \texttt{01 Oct 2026}, and Gemma receives text input only.

\vspace{0.1cm}
\noindent\textbf{Models and runtime.} Table~\ref{tab:checkpoint-config} lists the six checkpoints, and Table~\ref{tab:runtime-config} lists the runtime and decoding settings. All runs use one NVIDIA A40 GPU. The five BF16 models run in Transformers with deterministic algorithms, and Qwen3-32B-AWQ runs in vLLM with 4-bit weights and BF16 computation. Configuration files pin the checkpoint revisions, and run manifests record the full environment.

\begin{table}[t]
\centering
\small
\setlength{\tabcolsep}{7pt}
\setlength{\aboverulesep}{0pt}
\setlength{\belowrulesep}{0pt}
\renewcommand{\arraystretch}{1.2}
\begin{tabular}{lc}
\toprule
\textbf{Checkpoint} & \textbf{Weights}\\
\midrule
{\footnotesize\texttt{Qwen/Qwen3-8B}}                    & BF16\\
{\footnotesize\texttt{Qwen/Qwen3-14B}}                   & BF16\\
{\footnotesize\texttt{Qwen/Qwen3-32B-AWQ}}               & 4-bit AWQ\\
\hdashline[1pt/1.5pt]
{\footnotesize\texttt{meta-llama/Llama-3.1-8B-Instruct}} & BF16\\
{\footnotesize\texttt{google/gemma-3-12b-it}}            & BF16\\
{\footnotesize\texttt{allenai/OLMo-2-1124-13B-Instruct}} & BF16\\
\bottomrule
\end{tabular}
\caption{Model checkpoints on Hugging Face. BF16 checkpoints run in Transformers, and the AWQ checkpoint runs in vLLM with BF16 computation.}
\label{tab:checkpoint-config}
\end{table}

\begin{table}[t]
\centering
\small
\setlength{\tabcolsep}{4pt}
\setlength{\aboverulesep}{0pt}
\setlength{\belowrulesep}{0pt}
\renewcommand{\arraystretch}{1.25}
\begin{tabular}{>{\columncolor{gray!12}\bfseries}p{0.25\columnwidth}>{\raggedright\arraybackslash}p{0.68\columnwidth}@{}}
\toprule
Hardware      & One NVIDIA A40 GPU\\
\hdashline[1pt/1.5pt]
Framework     & PyTorch 2.7.0, CUDA 11.8, TF32 disabled\\
\hdashline[1pt/1.5pt]
BF16 models   & Transformers 4.57.6, SDPA math attention, deterministic algorithms\\
\hdashline[1pt/1.5pt]
AWQ model     & vLLM 0.9.2 (V0 engine), Transformers 4.53.2, AWQ Marlin kernels, FlashAttention, eager execution\\
\midrule
Batch size    & 1\\
\hdashline[1pt/1.5pt]
Seed          & 2027\\
\hdashline[1pt/1.5pt]
Input limit   & 4,096 tokens, longer inputs rejected\\
\hdashline[1pt/1.5pt]
Step 1        & Greedy over the letter tokens A--E\\
\hdashline[1pt/1.5pt]
Step 2        & End-of-sequence token only\\
\hdashline[1pt/1.5pt]
Budget        & 2 tokens\\
\bottomrule
\end{tabular}
\caption{Runtime environment (top) and answer decoding (bottom). Steps refer to the two decoding steps of each completion.}
\label{tab:runtime-config}
\vspace{-10pt}
\end{table}

\vspace{0.1cm}
\noindent\textbf{Constrained decoding.} Each request is processed alone, and inputs longer than 4,096 tokens are rejected without truncation. For answers, greedy selection at the first step is restricted to the canonical single-letter tokens for A--E in the pinned tokenizer, and only the model's end-of-sequence token is allowed at the second step. The model therefore returns the option letter with the highest probability, which is the prediction $\hat{o}$ in \S\ref{sec:conditions}. For scope judgments, greedy decoding is restricted to the token sequences of the three labels, each followed by the end-of-sequence token.

\vspace{0.1cm}
\noindent\textbf{Valid responses.} A response is valid if it contains a single letter from A--E and ends normally. Malformed or truncated outputs are classified as invalid and stay in every outcome-rate denominator. Raw completions and generated token IDs are saved with each run.

\subsection{Statistical Inference}
\label{app:metrics}

\noindent\textbf{Aggregation.} We use the evaluation subsets and metrics in \S\ref{sec:metrics}. Each model's rate or contrast is computed on its own subsets, and reported values are means over models with equal weight, so counts are never pooled across models. A model enters inference if its original $\mathcal A_m$ covers at least 20 distinct source clusters. All six models meet this rule, and the model set stays fixed during resampling.

\vspace{0.1cm}
\noindent\textbf{Cluster bootstrap.} We draw 10,000 replicates with seed 2027. Each replicate samples 489 source clusters with replacement from the 489 clusters of the main split, without stratification. Responses and the original membership of $\mathcal B_m$ and $\mathcal A_m$ stay fixed, and each rate is weighted by the cluster draws. For replicate $b$, the lure rate becomes
\[
\mathrm{LR}^{(b)}_{m,t} = \frac{\sum_{i \in \mathcal B_m} w^{(b)}_i\, \mathbb{I}\big[\hat{o}_{m,i}(t) = \ell_i\big]}{\sum_{i \in \mathcal B_m} w^{(b)}_i},
\]
where $w^{(b)}_i$ is the number of times the cluster of item $i$ is drawn. RLR and the other rates use the same weights, and contrasts and model means are then computed as in \S\ref{sec:inference}. The same draws apply to all models and conditions, so all comparisons remain paired.

\vspace{0.1cm}
\noindent\textbf{Undefined replicates.} A rate with an empty denominator is undefined, not zero. If a required rate is undefined for any model, the mean for that replicate is undefined, and no model is dropped from an individual replicate. Undefined replicates are excluded only from the intervals of the affected metric. All 10,000 replicates are valid for both primary contrasts.

\vspace{0.1cm}
\noindent\textbf{Confidence intervals.} We report two-sided percentile intervals with linear interpolation. The two primary contrasts, $\Delta_{\mathrm{LR}}$ and $\Delta_{\mathrm{RLR}}$, use Bonferroni-adjusted 97.5\% intervals, whose endpoints are the 1.25th and 98.75th percentiles. All other intervals are exploratory 95\% intervals, whose endpoints are the 2.5th and 97.5th percentiles.

\vspace{0.1cm}
\noindent\textbf{Other resampling units.} The sensitivity analysis repeats this procedure with the 132 dependency groups in Appendix~\ref{app:sources} as resampling units (Appendix~\ref{app:robustness}). Analyses on the 200-item diagnostic sample resample its 154 source clusters, and the option-order check also reports intervals over its 60 dependency groups. All other settings are unchanged, and all conditions, option orders, and context types of an item stay paired.

\section{Additional Results}
\label{app:results}

This section reports full outcome distributions and per-model results for \S\ref{sec:results} and \S\ref{sec:analysis}. Model names follow Table~\ref{tab:main-results}. All rates are percentages, means weight models equally, and differences are computed from unrounded rates. In tables, G$\to$S places G before S, and \emph{S with G} averages the two orders within each model.

\subsection{Full Outcome Distributions}
\label{app:outcomes}

Table~\ref{tab:extended-main-outcomes} gives the outcome distribution for every condition. Conditions without context and with G alone use all 959 items, context-alone conditions use $\mathcal B_m$, and conditions with G use $\mathcal A_m$. Context-alone rates are also reported on $\mathcal A_m$, so that comparisons before and after adding G use the same items. Table~\ref{tab:extended-order-results} gives per-model lure rates for each order on the same $\mathcal A_m$.

\begin{table}[h]
\centering
\small
\setlength{\tabcolsep}{5pt}
\setlength{\aboverulesep}{0pt}
\setlength{\belowrulesep}{0pt}
\renewcommand{\arraystretch}{1.2}
\begin{tabular}{llc>{\columncolor{gray!12}}ccc}
\toprule
\textbf{Subset} & \textbf{Condition} & \textbf{Gold} & \textbf{Lure} & \textbf{Abstain} & \textbf{Other}\\
\midrule
\multirow{2}{*}{All} & No context & 47.90 & 22.56 & 8.81 & 20.73\\
                     & G alone    & 98.68 & 0.57  & 0.16 & 0.59\\
\midrule
\multirow{2}{*}{$\mathcal B_m$} & S alone & 56.84 & 29.26 & 12.17 & 1.73\\
                                & F alone & 17.90 & 75.38 & 6.43  & 0.29\\
\midrule
\multirow{8}{*}{$\mathcal A_m$} & S alone & 56.90 & 29.23 & 12.18 & 1.69\\
                                & F alone & 17.95 & 75.38 & 6.43  & 0.25\\
\cdashline{2-6}[1pt/1.5pt]
 & G$\to$S & 99.38 & 0.46  & 0.03 & 0.12\\
 & S$\to$G & 98.26 & 1.36  & 0.04 & 0.34\\
 & G$\to$F & 95.24 & 3.91  & 0.77 & 0.08\\
 & F$\to$G & 58.91 & 38.10 & 2.71 & 0.27\\
\cdashline{2-6}[1pt/1.5pt]
 & S with G & 98.82 & 0.91  & 0.04 & 0.23\\
 & F with G & 77.08 & 21.01 & 1.74 & 0.18\\
\bottomrule
\end{tabular}
\caption{Outcome distributions on the main split, averaged equally over models. All denotes the 959 main items. Each row sums to 100 up to rounding.}
\label{tab:extended-main-outcomes}
\vspace{-10pt}
\end{table}

\begin{table}[h]
\centering
\small
\setlength{\tabcolsep}{6pt}
\setlength{\aboverulesep}{0pt}
\setlength{\belowrulesep}{0pt}
\renewcommand{\arraystretch}{1.2}
\begin{tabular}{lcc;{2pt/2pt}c>{\columncolor{gray!12}}c}
\toprule
& \multicolumn{2}{c;{2pt/2pt}}{\textbf{S with G}} & \multicolumn{2}{c}{\textbf{F with G}}\\
\cmidrule(lr){2-3}\cmidrule(lr){4-5}
\textbf{Model} & G first & S first & G first & F first\\
\midrule
Qwen3-8B      & 1.03 & 0.78 & 1.03 & 28.42\\
Qwen3-14B     & 0.22 & 1.10 & 0.44 & 51.54\\
Qwen3-32B-AWQ & 0.00 & 1.56 & 3.32 & 51.76\\
Llama-3.1-8B  & 0.00 & 0.22 & 5.62 & 30.79\\
Gemma-3-12B   & 0.00 & 1.21 & 5.87 & 35.83\\
OLMo-2-13B    & 1.53 & 3.27 & 7.19 & 30.28\\
\midrule
\textbf{Mean} & \textbf{0.46} & \textbf{1.36} & \textbf{3.91} & \textbf{38.10}\\
\bottomrule
\end{tabular}
\caption{Lure rates by order on $\mathcal A_m$. The order effect is the misleading-first rate minus the G-first rate.}
\label{tab:extended-order-results}
\vspace{-10pt}
\end{table}

\subsection{Paired Conditional Outcomes}
\label{app:paired}

This check pairs outcomes within each item. For each model and order, we keep the items in $\mathcal A_m$ answered correctly when S is paired with G, and compute outcomes on these items when F replaces S in the same position. The kept items cover 99.38\% of $\mathcal A_m$ when G comes first and 98.26\% when F comes first (Table~\ref{tab:paired-conditional-results}). The lure rates with F are 3.68\% (95\% CI: [2.93, 4.45]) and 37.32\% ([33.01, 41.40]), close to the rates of 3.91\% and 38.10\% on all of $\mathcal A_m$. Abstention rates are 0.74\% and 2.71\%, and other distractors account for 0.08\% and 0.28\%. The kept items can differ between orders, so the difference between the two rates is not an order effect on a fixed subset.

\begin{table}[t]
\centering
\small
\setlength{\tabcolsep}{12pt}
\setlength{\aboverulesep}{0pt}
\setlength{\belowrulesep}{0pt}
\renewcommand{\arraystretch}{1.2}
\begin{tabular}{lcc>{\columncolor{gray!12}}c}
\toprule
\textbf{Model} & $n$ & \textbf{Gold} & \textbf{Lure}\\
\midrule
\multicolumn{4}{l}{\textit{G first}}\\
Qwen3-8B      & 381 & 96.85 & 0.79\\
Qwen3-14B     & 453 & 99.56 & 0.22\\
Qwen3-32B-AWQ & 512 & 96.68 & 3.32\\
Llama-3.1-8B  & 445 & 93.48 & 5.62\\
Gemma-3-12B   & 493 & 92.90 & 5.68\\
OLMo-2-13B    & 451 & 93.57 & 6.43\\
\hdashline[1pt/1.5pt]
\textbf{Mean} & & \textbf{95.51} & \textbf{3.68}\\
\midrule
\multicolumn{4}{l}{\textit{F first}}\\
Qwen3-8B      & 382 & 63.09 & 27.49\\
Qwen3-14B     & 447 & 44.52 & 51.01\\
Qwen3-32B-AWQ & 504 & 48.61 & 50.99\\
Llama-3.1-8B  & 441 & 68.48 & 30.61\\
Gemma-3-12B   & 488 & 62.70 & 35.25\\
OLMo-2-13B    & 441 & 70.75 & 28.57\\
\hdashline[1pt/1.5pt]
\textbf{Mean} & & \textbf{59.69} & \textbf{37.32}\\
\bottomrule
\end{tabular}
\caption{Outcomes with F and G on items in $\mathcal A_m$ answered correctly with S and G in the same order. $n$ is the number of kept items per model, and abstention and other distractors stay in the denominator.}
\label{tab:paired-conditional-results}
\vspace{-10pt}
\end{table}

\subsection{Condition Supplement}
\label{app:condition}

We evaluate the 25 condition-supplement items with the same conditions, prompts, models, and scoring as the main split, and analyze them separately. Subsets $\mathcal B_m$ and $\mathcal A_m$ follow the same rules and coincide for every model. Table~\ref{tab:supplement-model-results} gives per-model lure rates, and Table~\ref{tab:supplement-outcomes} gives outcome distributions. With 10--15 items per model, these results are descriptive. As in the main split, F misleads more than S, and adding G removes the effect of S.

\begin{table}[h]
\centering
\small
\setlength{\tabcolsep}{7pt}
\setlength{\aboverulesep}{0pt}
\setlength{\belowrulesep}{0pt}
\renewcommand{\arraystretch}{1.2}
\begin{tabular}{lc;{2pt/2pt}cc;{2pt/2pt}c}
\toprule
& & \multicolumn{2}{c;{2pt/2pt}}{\textbf{Alone}} & \textbf{With G}\\
\cmidrule(lr){3-4}\cmidrule(lr){5-5}
\textbf{Model} & $n$ & $\mathrm{LR}_{\mathrm S}$ & $\mathrm{LR}_{\mathrm F}$ & $\mathrm{RLR}_{\mathrm F}$\\
\midrule
Qwen3-8B      & 13 & 15.38 & 46.15 & 3.85\\
Qwen3-14B     & 15 & 20.00 & 60.00 & 10.00\\
Qwen3-32B-AWQ & 10 & 20.00 & 80.00 & 10.00\\
Llama-3.1-8B  & 15 & 20.00 & 73.33 & 6.67\\
Gemma-3-12B   & 11 & 9.09  & 72.73 & 4.55\\
OLMo-2-13B    & 12 & 41.67 & 58.33 & 0.00\\
\midrule
\textbf{Mean} & & \textbf{21.02} & \textbf{65.09} & \textbf{5.84}\\
\bottomrule
\end{tabular}
\caption{Lure rates on the condition supplement, where $n = |\mathcal B_m| = |\mathcal A_m|$. RLR averages both orders, and $\mathrm{RLR}_{\mathrm S}$ is 0.00 for every model.}
\label{tab:supplement-model-results}
\vspace{-10pt}
\end{table}

\begin{table}[h]
\centering
\small
\setlength{\tabcolsep}{4.5pt}
\setlength{\aboverulesep}{0pt}
\setlength{\belowrulesep}{0pt}
\renewcommand{\arraystretch}{1.2}
\begin{tabular}{llc>{\columncolor{gray!12}}ccc}
\toprule
\textbf{Subset} & \textbf{Condition} & \textbf{Gold} & \textbf{Lure} & \textbf{Abstain} & \textbf{Other}\\
\midrule
\multirow{2}{*}{All} & No context & 50.67 & 18.67 & 1.33 & 29.33\\
                     & G alone    & 96.00 & 1.33  & 0.67 & 2.00\\
\midrule
\multirow{2}{*}{$\mathcal B_m$} & S alone & 74.36 & 21.02 & 4.62 & 0.00\\
                                & F alone & 22.82 & 65.09 & 3.91 & 8.18\\
\midrule
\multirow{2}{*}{$\mathcal A_m$} & S with G & 100.00 & 0.00 & 0.00 & 0.00\\
                                & F with G & 90.93  & 5.84 & 2.59 & 0.64\\
\bottomrule
\end{tabular}
\caption{Outcome distributions on the condition supplement. All denotes the 25 supplement items. Values are percentages, and rows with G average both orders.}
\label{tab:supplement-outcomes}
\vspace{-10pt}
\end{table}

\subsection{Scope Recognition}
\label{app:scope}

\noindent\textbf{Sample.} We sample 40 items without replacement from each of the five source--scope groups, using seed 2027 and no model responses. The 200 items span 154 source clusters and are also used for the option-order check (Appendix~\ref{app:robustness}).

\vspace{0.1cm}
\noindent\textbf{Judgments.} Each model judges G, S, and F for every sampled item with the prompt in Figure~\ref{fig:prompt-analysis}(a). The correct label is \emph{Applicable} for G and F and \emph{Not applicable} for S, since the false value in F does not change its scope. \emph{Cannot determine} counts as incorrect and stays in the denominator. Every model judges all 200 G contexts applicable, and Table~\ref{tab:scope-label-counts} gives the label counts for F and S.

\begin{table}[h]
\centering
\small
\setlength{\tabcolsep}{5.7pt}
\setlength{\aboverulesep}{0pt}
\setlength{\belowrulesep}{0pt}
\renewcommand{\arraystretch}{1.2}
\begin{tabular}{l>{\columncolor{gray!12}}ccc;{2pt/2pt}c>{\columncolor{gray!12}}cc}
\toprule
& \multicolumn{3}{c;{2pt/2pt}}{\textbf{F}} & \multicolumn{3}{c}{\textbf{S}}\\
\cmidrule(lr){2-4}\cmidrule(lr){5-7}
\textbf{Model} & A & N & C & A & N & C\\
\midrule
Qwen3-8B      & 194 & 6  & 0 & 88  & 112 & 0\\
Qwen3-14B     & 187 & 13 & 0 & 21  & 179 & 0\\
Qwen3-32B-AWQ & 198 & 2  & 0 & 109 & 91  & 0\\
Llama-3.1-8B  & 191 & 5  & 4 & 88  & 65  & 47\\
Gemma-3-12B   & 192 & 8  & 0 & 6   & 194 & 0\\
OLMo-2-13B    & 190 & 10 & 0 & 89  & 108 & 3\\
\midrule
\textbf{Total} & \textbf{1152} & \textbf{44} & \textbf{4} & \textbf{401} & \textbf{749} & \textbf{50}\\
\bottomrule
\end{tabular}
\caption{Scope labels for F and S out of 200 items per model. A, N, and C denote \emph{Applicable}, \emph{Not applicable}, and \emph{Cannot determine}, and shaded columns mark the correct label. Every G label is \emph{Applicable}.}
\label{tab:scope-label-counts}
\vspace{-10pt}
\end{table}

\vspace{0.1cm}
\noindent\textbf{Judgments and answers.} We link each S judgment to the same model's original answers without context and with S alone. Let $\mathcal H_m$ be the sampled items in $\mathcal B_m$. For each label, the conditional lure rate is the share of items in $\mathcal H_m$ with that label on which the model chooses the lure with S alone, and the mean averages these model-level rates. Table~\ref{tab:scope-answer-counts} gives the counts. The mean rate is 20.34\% for items judged \emph{Not applicable} and 66.06\% for items judged \emph{Applicable}, although the latter includes small subsets, such as one item for Gemma-3-12B. Judging and answering are separate requests, so these rates describe an association.

\begin{table}[t]
\centering
\small
\setlength{\tabcolsep}{6.7pt}
\setlength{\aboverulesep}{0pt}
\setlength{\belowrulesep}{0pt}
\renewcommand{\arraystretch}{1.2}
\begin{tabular}{lc;{2pt/2pt}c>{\columncolor{gray!12}}cc}
\toprule
& & \multicolumn{3}{c}{\textbf{S judged as}}\\
\cmidrule(lr){3-5}
\textbf{Model} & \textbf{All} & A & N & C\\
\midrule
Qwen3-8B      & 16/80  & 12/22 & 4/58  & ---\\
Qwen3-14B     & 15/97  & 5/6   & 10/91 & ---\\
Qwen3-32B-AWQ & 21/101 & 16/51 & 5/50  & ---\\
Llama-3.1-8B  & 19/85  & 10/30 & 6/32  & 3/23\\
Gemma-3-12B   & 14/100 & 1/1   & 13/99 & ---\\
OLMo-2-13B    & 69/94  & 30/32 & 38/61 & 1/1\\
\midrule
\textbf{Mean (\%)} & \textbf{27.67} & \textbf{66.06} & \textbf{20.34} & ---\\
\bottomrule
\end{tabular}
\caption{Lure counts with S alone over item counts in $\mathcal H_m$, the sampled items in $\mathcal B_m$, overall and by the model's S label. A dash marks an empty subset, and the mean averages model-level rates.}
\label{tab:scope-answer-counts}
\vspace{-10pt}
\end{table}

\subsection{Context-Faithful Instruction}
\label{app:faithful}

\noindent\textbf{Setup.} We rerun all 959 items under the seven conditions with context, using the CF prompt (Figure~\ref{fig:prompt-analysis}b). Models, chat templates, decoding, questions, options, contexts, and their orders are unchanged, and no scope judgments enter the inputs. As a consistency check, we also rerun the no-context condition with the original answering prompt, giving 46,032 responses in total. The no-context answers match the original run for all 5,754 item--model pairs. Comparisons use the original $\mathcal B_m$ and $\mathcal A_m$, except that G alone uses all 959 items. Gold and lure keep their original meaning. CF prescribes the lure for F alone, and abstention for S alone and for F with G.

\vspace{0.1cm}
\noindent\textbf{Results.} Table~\ref{tab:cf-complete-outcomes} compares outcome distributions under the two prompts, and Table~\ref{tab:cf-model-results} gives per-model rates. These comparisons are descriptive.

\begin{table}[h]
\centering
\small
\setlength{\tabcolsep}{4.5pt}
\setlength{\aboverulesep}{0pt}
\setlength{\belowrulesep}{0pt}
\renewcommand{\arraystretch}{1.2}
\begin{tabular}{llcccc}
\toprule
\textbf{Condition} & \textbf{Prompt} & \textbf{Gold} & \textbf{Lure} & \textbf{Abstain} & \textbf{Other}\\
\midrule
                                          & Original & 98.68 & 0.57 & 0.16 & 0.59\\
\rowcolor{gray!12}\multirow{-2}{*}{\cellcolor{white}G alone} & CF & \textbf{91.26} & 3.11 & 0.80 & 4.83\\
\hdashline[1pt/1.5pt]
                                          & Original & 56.84 & 29.26 & 12.17 & 1.73\\
\rowcolor{gray!12}\multirow{-2}{*}{\cellcolor{white}S alone} & CF & 6.99 & 15.06 & \textbf{77.85} & 0.10\\
\hdashline[1pt/1.5pt]
                                          & Original & 17.90 & 75.38 & 6.43 & 0.29\\
\rowcolor{gray!12}\multirow{-2}{*}{\cellcolor{white}F alone} & CF & 13.92 & \textbf{69.43} & 13.18 & 3.47\\
\hdashline[1pt/1.5pt]
                                          & Original & 98.82 & 0.91 & 0.04 & 0.23\\
\rowcolor{gray!12}\multirow{-2}{*}{\cellcolor{white}S with G} & CF & \textbf{93.84} & 1.88 & 2.02 & 2.26\\
\hdashline[1pt/1.5pt]
                                          & Original & 77.08 & 21.01 & 1.74 & 0.18\\
\rowcolor{gray!12}\multirow{-2}{*}{\cellcolor{white}F with G} & CF & 42.26 & 12.59 & \textbf{44.48} & 0.68\\
\bottomrule
\end{tabular}
\caption{Outcome distributions under the original and CF prompts. Bold marks the response CF prescribes. G alone uses all 959 items, context-alone conditions use $\mathcal B_m$, and conditions with G use $\mathcal A_m$.}
\label{tab:cf-complete-outcomes}
\vspace{-10pt}
\end{table}

\begin{table}[t]
\centering
\small
\setlength{\tabcolsep}{6.4pt}
\setlength{\aboverulesep}{0pt}
\setlength{\belowrulesep}{0pt}
\renewcommand{\arraystretch}{1.2}
\begin{tabular}{lccc}
\toprule
& \multicolumn{2}{c}{\textbf{S alone}} & \textbf{F alone}\\
\cmidrule(lr){2-3}\cmidrule(lr){4-4}
\textbf{Model} & Abstain & Lure & Lure\\
\midrule
Qwen3-8B      & 86.86 & 5.93  & 43.30\\
Qwen3-14B     & 87.00 & 2.64  & 48.90\\
Qwen3-32B-AWQ & 78.32 & 9.57  & 87.70\\
Llama-3.1-8B  & 93.71 & 2.47  & 65.62\\
Gemma-3-12B   & 92.51 & 7.09  & 93.52\\
OLMo-2-13B    & 28.73 & 62.63 & 77.54\\
\hdashline[1pt/1.5pt]
\textbf{Mean} & \textbf{77.85} & \textbf{15.06} & \textbf{69.43}\\
\midrule
& \textbf{G alone} & \textbf{S with G} & \textbf{F with G}\\
\cmidrule(lr){2-2}\cmidrule(lr){3-3}\cmidrule(lr){4-4}
\textbf{Model} & Gold & Gold & Abstain\\
\midrule
Qwen3-8B      & 77.89 & 95.09 & 91.99\\
Qwen3-14B     & 81.33 & 92.18 & 35.24\\
Qwen3-32B-AWQ & 99.37 & 96.09 & 7.23\\
Llama-3.1-8B  & 93.01 & 95.73 & 11.24\\
Gemma-3-12B   & 98.96 & 94.64 & 99.39\\
OLMo-2-13B    & 96.98 & 89.32 & 21.79\\
\hdashline[1pt/1.5pt]
\textbf{Mean} & \textbf{91.26} & \textbf{93.84} & \textbf{44.48}\\
\bottomrule
\end{tabular}
\caption{Per-model rates under CF for the outcomes discussed in \S\ref{sec:instruction}. Subsets follow Table~\ref{tab:cf-complete-outcomes}.}
\label{tab:cf-model-results}
\end{table}

\vspace{0.1cm}
\noindent\textbf{Common subsets.} As a check, we restrict each $\mathcal A_m$ to the items the model also answers correctly with G alone under CF:
\[
\mathcal A_m^{\mathrm{common}} = \big\{\, i \in \mathcal A_m : \hat{o}^{\mathrm{CF}}_{m,i}(\mathrm G) = y^*_i \,\big\},
\]
where $\hat{o}^{\mathrm{CF}}$ denotes predictions under CF. These subsets keep 97.86\% of $\mathcal A_m$ on average (Table~\ref{tab:cf-common-subsets}). On them, the gold rate with S and G is 98.92\% under the original prompt and 94.82\% under CF. CF changes context use, reliance on outside knowledge, and abstention together, so these results do not isolate the effect of any single clause.

\begin{table}[t]
\centering
\small
\setlength{\tabcolsep}{8.5pt}
\setlength{\aboverulesep}{0pt}
\setlength{\belowrulesep}{0pt}
\renewcommand{\arraystretch}{1.2}
\begin{tabular}{lcc>{\columncolor{gray!12}}c}
\toprule
\textbf{Model} & $|\mathcal A_m|$ & $|\mathcal A_m^{\mathrm{common}}|$ & \textbf{Kept}\\
\midrule
Qwen3-8B      & 387 & 368 & 95.09\\
Qwen3-14B     & 454 & 436 & 96.04\\
Qwen3-32B-AWQ & 512 & 511 & 99.80\\
Llama-3.1-8B  & 445 & 433 & 97.30\\
Gemma-3-12B   & 494 & 493 & 99.80\\
OLMo-2-13B    & 459 & 455 & 99.13\\
\midrule
\textbf{Mean} & & & \textbf{97.86}\\
\bottomrule
\end{tabular}
\caption{Original and common subset sizes for the CF check. Kept is the share of $\mathcal A_m$ retained.}
\label{tab:cf-common-subsets}
\vspace{-10pt}
\end{table}

\subsection{Robustness Checks}
\label{app:robustness}

\noindent\textbf{Model composition.} Excluding OLMo-2-13B and averaging over the other five models, with their subsets and responses unchanged, lowers the lure rate with S alone from 29.26\% to 20.55\%. The F--S differences rise to 54.57 points for LR and 20.85 points for RLR.

\vspace{0.1cm}
\noindent\textbf{Source--scope groups.} We restrict each model's $\mathcal B_m$ and $\mathcal A_m$ to the items of each source--scope group and compute rates per model. Table~\ref{tab:source-scope-rates} gives the means, and Table~\ref{tab:source-scope-denominators} gives the subset sizes. These group rates are not reweighted to the full main split. Sources and scope dimensions are only partly crossed, so group differences do not separate scope dimension from source content.

\begin{table}[t]
\centering
\small
\setlength{\tabcolsep}{5.5pt}
\setlength{\aboverulesep}{0pt}
\setlength{\belowrulesep}{0pt}
\renewcommand{\arraystretch}{1.2}
\begin{tabular}{lcc;{2pt/2pt}cc}
\toprule
& \multicolumn{2}{c;{2pt/2pt}}{\textbf{Alone}} & \multicolumn{2}{c}{\textbf{With G}}\\
\cmidrule(lr){2-3}\cmidrule(lr){4-5}
\textbf{Group} & $\mathrm{LR}_{\mathrm S}$ & $\mathrm{LR}_{\mathrm F}$ & $\mathrm{RLR}_{\mathrm S}$ & $\mathrm{RLR}_{\mathrm F}$\\
\midrule
FRED time           & 35.91 & 98.48 & 0.42 & 36.05\\
FRED location       & 30.97 & 99.57 & 0.21 & 33.86\\
\hdashline[1pt/1.5pt]
SituatedQA time     & 28.00 & 82.52 & 0.65 & 18.04\\
SituatedQA location & 7.59  & 37.32 & 0.17 & 4.15\\
\hdashline[1pt/1.5pt]
AmbigDocs entity    & 73.62 & 94.74 & 4.94 & 29.83\\
\bottomrule
\end{tabular}
\caption{Lure rates within source--scope groups. LR uses each group's part of $\mathcal B_m$, and RLR uses its part of $\mathcal A_m$ and averages both orders.}
\label{tab:source-scope-rates}
\end{table}

\begin{table}[t]
\centering
\small
\setlength{\tabcolsep}{7pt}
\setlength{\aboverulesep}{0pt}
\setlength{\belowrulesep}{0pt}
\renewcommand{\arraystretch}{1.2}
\begin{tabular}{lcc;{2pt/2pt}cc;{2pt/2pt}c}
\toprule
\textbf{Model} & \textbf{FT} & \textbf{FL} & \textbf{ST} & \textbf{SL} & \textbf{AE}\\
\midrule
Qwen3-8B      & 73 & 66 & 74  & 133 & 42\\
Qwen3-14B     & 88 & 72 & 95  & 149 & 50\\
Qwen3-32B-AWQ & 95 & 89 & 99  & 156 & 73\\
Llama-3.1-8B  & 89 & 66 & 88  & 148 & 54\\
Gemma-3-12B   & 92 & 75 & 110 & 152 & 65\\
OLMo-2-13B    & 89 & 96 & 74  & 141 & 63\\
\bottomrule
\end{tabular}
\caption{Sizes of $\mathcal B_m$ within source--scope groups: FRED time (FT) and location (FL), SituatedQA time (ST) and location (SL), and AmbigDocs entity (AE). $\mathcal A_m$ has the same size except for Qwen3-8B on ST (73) and OLMo-2-13B on FT (88), SL (140), and AE (61).}
\label{tab:source-scope-denominators}
\vspace{-10pt}
\end{table}

\vspace{0.1cm}
\noindent\textbf{Dependency-group resampling.} Table~\ref{tab:dependency-intervals} compares intervals from source-cluster and dependency-group resampling (Appendix~\ref{app:metrics}). Point estimates are unchanged. Both primary F--S intervals and the interval for the order effect of F stay above zero, while the interval for the order effect of S includes zero.

\begin{table}[h]
\centering
\small
\setlength{\tabcolsep}{5.5pt}
\setlength{\aboverulesep}{0pt}
\setlength{\belowrulesep}{0pt}
\renewcommand{\arraystretch}{1.4}
\begin{tabular}{lc;{2pt/2pt}c>{\columncolor{gray!12}}c}
\toprule
& & \multicolumn{2}{c}{\textbf{Interval}}\\
\cmidrule(lr){3-4}
\textbf{Metric} & \textbf{Est.} & \textbf{Clusters} & \textbf{Groups}\\
\midrule
$\Delta_{\mathrm{LR}}^{\dagger}$  & 46.12 & [41.86, 49.95] & [30.03, 57.11]\\
$\Delta_{\mathrm{RLR}}^{\dagger}$ & 20.10 & [17.55, 22.56] & [9.13, 28.11]\\
\hdashline[1pt/1.5pt]
F reduction     & 54.37 & [51.45, 57.13] & [44.01, 60.27]\\
S reduction     & 28.32 & [25.92, 30.76] & [18.88, 34.59]\\
\hdashline[1pt/1.5pt]
F order effect  & 34.19 & [30.20, 38.02] & [14.93, 47.04]\\
S order effect  & 0.89  & [0.33, 1.55]   & [$-$0.04, 3.03]\\
\bottomrule
\end{tabular}
\caption{Estimates and intervals in percentage points under source-cluster and dependency-group resampling. $^\dagger$ marks the primary contrasts, which use 97.5\% intervals, and other rows use 95\%. Reduction is the context-alone rate minus RLR on $\mathcal A_m$.}
\label{tab:dependency-intervals}
\vspace{-10pt}
\end{table}

\vspace{0.1cm}
\noindent\textbf{Option order.} On the 200-item diagnostic sample, we arrange the original A--D option texts $(a,b,c,d)$ in three orders,
\[
(a,b,c,d),\qquad (b,c,d,a),\qquad (c,d,a,b),
\]
with E fixed and the gold and lure keys following their texts. Each model answers every sampled item without context, with S alone, and with F alone under each order, giving 10,800 responses, and answers under the original order match the main evaluation. Let $\mathcal B_m^{(r)}$ contain the sampled items that model $m$ answers correctly without context under order $r \in \{0,1,2\}$, where $r=0$ is the original order, and let
\[
\mathcal B_m^{\mathrm{stable}} = \bigcap_{r=0}^{2} \mathcal B_m^{(r)}.
\]
Eligibility uses only answers without context, and $\mathcal B_m^{\mathrm{stable}}$ keeps 62.22\% of $\mathcal B_m^{(0)}$ on average. Table~\ref{tab:option-robustness} reports lure rates for the original order on $\mathcal B_m^{(0)}$, the original order on $\mathcal B_m^{\mathrm{stable}}$, and the mean over the three orders on $\mathcal B_m^{\mathrm{stable}}$. The first two settings hold the order fixed, and the last two hold the items fixed. The stricter rule also changes which items are kept, so lower rates on $\mathcal B_m^{\mathrm{stable}}$ cannot be attributed to option position alone.

\begin{figure*}[!t]
{\small\textbf{(a)} Scope judgment}\par\vspace{3pt}
\begin{lstlisting}[style=prompt, backgroundcolor={}]
Determine whether the assertion in the material refers to the same target scope as the question, including the entity, time, and location specified in the question. Ignore whether the answer value is factually correct; judge only whether the scope matches. Output exactly one label: Applicable, Not applicable, or Cannot determine.

Question:
(*\promptfield{question}*)

Options:
(*\promptfield{options}*)

Material:
(*\promptfield{context}*)
\end{lstlisting}
\vspace{8pt}
{\small\textbf{(b)} Context-faithful answering}\par\vspace{3pt}
\begin{lstlisting}[style=prompt, backgroundcolor={}]
Answer the following multiple-choice question using only the supplied evidence that applies to the exact entity, time, population, and conditions asked about. Do not use outside knowledge to correct claims in the supplied evidence. Ignore evidence about a different scope. Choose E if the applicable evidence is insufficient to determine an answer, or if applicable passages support conflicting answers that cannot be resolved from the supplied evidence. Return only A, B, C, D, or E.

Question:
(*\promptfield{question}*)

Options:
(*\promptfield{options}*)

(*\promptfield{context 1}*)

(*\promptfield{context 2}*)
\end{lstlisting}
\caption{Prompts for the analyses in \S\ref{sec:analysis}, reproduced verbatim. (a) Scope judgment, with one context per request. (b) Context-faithful answering, with fields filled as in Figure~\ref{fig:prompt}.}
\label{fig:prompt-analysis}
\end{figure*}

\end{document}